\documentclass[10pt,twocolumn,letterpaper]{article}

\usepackage{cvpr}              %

\usepackage[utf8]{inputenc} %
\usepackage[T1]{fontenc}    %
\usepackage{amsmath}
\usepackage{amsfonts}
\usepackage{amssymb}
\usepackage{url}
\usepackage{graphicx}
\usepackage{epigraph}
\usepackage{bm}
\usepackage{bbm}
\usepackage[most]{tcolorbox}
\usepackage{capt-of}
\usepackage{caption}
\usepackage{subcaption}
\usepackage{colortbl}
\usepackage{wrapfig}
\usepackage{booktabs}
\usepackage{multirow}
\usepackage[table,dvipsnames]{xcolor}
\usepackage{pifont}
\usepackage{adjustbox}    
\usepackage{float}
\usepackage{tipa}
\usepackage{cuted}
\usepackage{tabu}
\usepackage[shortlabels]{enumitem}
\usepackage[toc,page,header,title]{appendix}
\usepackage{minitoc}
\usepackage{xspace}
\usepackage[pagebackref,breaklinks,colorlinks,citecolor=blue, linkcolor=blue]{hyperref}
\usepackage{mathrsfs}
\usepackage{wrapfig}

\usepackage{tikz}
\usepackage{tikz-3dplot}

\usepackage{algorithm}
\usepackage{algpseudocode}

\usepackage{etoc}
\usepackage{titletoc}

\makeatletter
\renewcommand\paragraph{\@startsection{paragraph}{4}{\z@}{0.6ex}{-1em}{\normalfont\normalsize\bfseries}}
\makeatother
\newcommand{\midsepremove}{\aboverulesep = 0.2mm \belowrulesep = 0.3mm}
\midsepremove
\algrenewcommand\algorithmiccomment[1]{\hfill\(\triangleright\)\ \textcolor{ForestGreen}{#1}}
\def\paperID{****} %
\def\confName{CVPR}
\def\confYear{2026}

\definecolor{forestgreen}{RGB}{34,139,34}

\tcbset{
  myquestionbox/.style={
    colback=forestgreen!5!white,
    colframe=forestgreen!60!black,
    coltitle=black,
    fonttitle=\normalfont\bfseries,
    enhanced,
    attach boxed title to top left={yshift=-2mm, xshift=5mm},
    boxed title style={
      colback=forestgreen!20!white,
      colframe=forestgreen!30!black,
      boxrule=0.4pt,
      rounded corners, 
    },
    rounded corners, 
    boxrule=0.8pt,
    width=\linewidth,
  }
}

\newcommand{\alert}[1]{{\color{black}{#1}}}

\newcommand{\citeme}[1]{\alert{[X]}\xspace}
\newcommand{\refme}[1]{\alert{(X)}}

\newcommand{\cmark}{\ding{51}}%
\newcommand{\xmark}{\ding{55}}%

\newcommand{\cmarkc}{{\color{forestgreen}\cmark}}

\newcommand{\xmarkc}{{\color{red}\xmark}}
\newcommand{\umarkc}{{\color{orange}\textbf{?}}}

\newcommand{\methodname}{\texttt{Franca}\xspace}
\definecolor{rowmint}{RGB}{220, 255, 240}

\title{\methodname: Nested Matryoshka Clustering for Scalable Visual Representation Learning}
\author{%
  Shashanka Venkataramanan$^{1\dagger}$\thanks{work done in Valeo.ai}
  \and
  Valentinos Pariza$^{2}$\hspace{1pt}\thanks{equal  contribution.}
  \and
  Mohammadreza Salehi$^{2,3}$
  \and
  Lukas Knobel$^{2}$
  \and
  Elias Ramzi$^{1}$
  \and
  Spyros Gidaris$^1$
  \and
  Andrei Bursuc$^1$\thanks{equal last authors. \\ Correspondence: andrei.bursuc@valeo.com} %
  \and
  Yuki M. Asano$^{2\hspace{1pt}\ddagger}$ \\\\
   $^1$Qualcomm AI Research $^2$ Fundamental AI Lab, UTN $^3$ VIS Lab, UvA  \\[1em]
}

\begin{document}
\maketitle
\setcounter{page}{1}
\begin{strip}
    \vspace{-42pt}
    \centering
    \includegraphics[width=\linewidth]{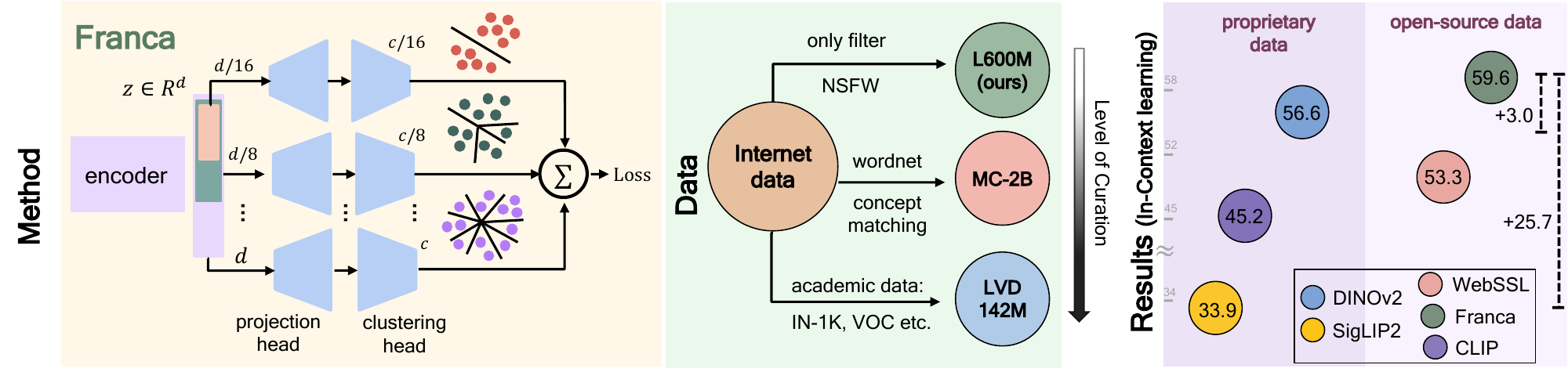}
    \captionof{figure}{\emph{Overview of~\methodname}. Left: We learn efficient \emph{Matryoshka-style}~\citep{kusupati2022matryoshka} visual representations using a multi-head clustering projection head. The encoder produces features $z \in \mathbb{R}^d$, which are sliced into progressively smaller subsets of dimensions $d,\dots d/8, d/16$. Each slice passes through a projection head and a corresponding clustering head with cluster counts $c,\dots, c/8, c/16$, inducing a \emph{coarse-to-fine hierarchy} of semantic abstraction. Middle: Unlike prior methods trained on proprietary data like WebLI in SigLIP 2 or curated academic datasets, e.g., LVD-142M in DINOv2,~\methodname is trained on open-source internet-scale \emph{minimally}-curated data. Right: It generalizes to dense tasks outperforming models trained on proprietary data. 
    }
    \label{fig:teaser}
    \vspace{-1pt}
\end{strip}

\begin{abstract}
 \label{sec:abstract}
\vspace{-1pt}
We present \emph{~\methodname} (pronounced \textit{Fran-ka}): \textit{`free' one}; the first fully open-source (data, code, weights) vision foundation model that matches—and in many cases surpasses—the performance of state-of-the-art proprietary models, e.g., DINOv2, CLIP, SigLIPv2, etc. Our approach is grounded in a transparent training pipeline inspired from Web-SSL and uses publicly available data: Imagenet-21K and LAION-COCO (600M images). Beyond model release, we tackle critical limitations in 
self-supervised learning  clustering methods. Existing approaches assign image features to large codebooks via clustering algorithms such as Sinkhorn-Knopp, but they often overlook the inherent ambiguity in cluster semantics. To address this, we introduce a multi-head clustering projector based on nested Matryoshka representations. This design progressively refines features into increasingly fine-grained clusters without increasing the model size, producing higher-quality dense representations. Additionally, we propose a novel positional disentanglement strategy that explicitly removes positional biases from dense representations.
This leads to consistent gains on several downstream tasks, demonstrating the utility of cleaner feature spaces.

\end{abstract}

\vspace{-12pt}
\section{Introduction}

Self-supervised learning (SSL) offers a scalable approach to training Vision Foundation Models (VFMs) by leveraging the abundance of image-only data, which far exceeds the availability of paired image-caption data. This enables the learning of highly generalizable visual representations.

\begin{table*}[t] %
\begin{center}
\setlength{\tabcolsep}{3.0pt}
\footnotesize
\begin{tabular}{lccccccccc}
\toprule
\textsc{Attribute / Model} & \textsc{MetaCLIP} & \textsc{Web-SSL} & \textsc{SigLIP 2} & \textsc{AIMv2} & \textsc{CLIP} & \textsc{DINOv2} & \textsc{RADIOv2.5} & \textsc{OpenCLIP} & \textsc{\methodname} \\
\midrule
\midrule
\multicolumn{10}{c}{\textbf{Training Code/ Checkpoints}} \\
\midrule
Model Publicly Available?         & \cellcolor{green!20}\cmarkc & \cellcolor{green!20}\cmarkc & \cellcolor{green!20}\cmarkc & \cellcolor{green!20}\cmarkc & \cellcolor{green!20}\cmarkc & \cellcolor{green!20}\cmarkc & \cellcolor{green!20}\cmarkc & \cellcolor{green!20}\cmarkc & \cellcolor{green!20}\cmarkc \\
Training Code Public?             & \cellcolor{green!20}\cmarkc & \cellcolor{green!20}\cmarkc & \cellcolor{red!20}\xmarkc & \cellcolor{red!20}\xmarkc & \cellcolor{red!20}\xmarkc & \cellcolor{green!20}\cmarkc & \cellcolor{red!20}\xmarkc & \cellcolor{green!20}\cmarkc & \cellcolor{green!20}\cmarkc \\
Intermediate Weights Public?     & \cellcolor{red!20}\xmarkc & \cellcolor{red!20}\xmarkc & \cellcolor{red!20}\xmarkc & \cellcolor{red!20}\xmarkc & \cellcolor{red!20}\xmarkc & \cellcolor{red!20}\xmarkc & \cellcolor{red!20}\xmarkc & \cellcolor{green!20}\cmarkc & \cellcolor{green!20}\cmarkc \\
\midrule
\multicolumn{10}{c}{\textbf{Training Data}} \\
\midrule
Training Data Public?             & \cellcolor{green!20}\cmarkc & \cellcolor{green!20}\cmarkc & \cellcolor{red!20}\xmarkc & \cellcolor{red!20}\xmarkc & \cellcolor{red!20}\xmarkc & \cellcolor{red!20}\xmarkc &  \cellcolor{red!20}\xmarkc & \cellcolor{green!20}\cmarkc & \cellcolor{green!20}\cmarkc \\
Training Data Downloadable?             & \cellcolor{red!20}\xmarkc & \cellcolor{red!20}\xmarkc & \cellcolor{red!20}\xmarkc & \cellcolor{red!20}\xmarkc & \cellcolor{red!20}\xmarkc & \cellcolor{red!20}\xmarkc &  \cellcolor{red!20}\xmarkc & \cellcolor{green!20}\cmarkc & \cellcolor{green!20}\cmarkc \\
Data Deduplication Code Public?  & \cellcolor{red!20}\xmarkc & \cellcolor{red!20}\xmarkc & \cellcolor{red!20}\xmarkc & \cellcolor{red!20}\xmarkc & \cellcolor{red!20}\xmarkc & \cellcolor{red!20}\xmarkc &  \cellcolor{red!20}\xmarkc & \cellcolor{green!20}\cmarkc & \cellcolor{green!20}\cmarkc \\
Data NSFW \& CSAM Filtered?     & \cellcolor{red!20}\xmarkc & \cellcolor{red!20}\xmarkc & \cellcolor{green!20}\cmarkc & \cellcolor{yellow!20}\umarkc & \cellcolor{yellow!20}\umarkc & \cellcolor{yellow!20}\umarkc & \cellcolor{yellow!20}\umarkc &  \cellcolor{green!20}\cmarkc  & \cellcolor{green!20}\cmarkc \\
\bottomrule
\end{tabular}
\vspace{-6pt}
\caption{\emph{Openness of Visual Foundation Models.} We analyze various models based on the public availability of their components MetaCLIP~\citep{xu2024demystifying}, Web-SSL~\citep{fan2025scaling}, SigLIP 2~\citep{tschannen2025siglip2}, AIMv2~\citep{fini2025multimodal}, CLIP~\citep{radford2021learning}, DINOv2~\citep{oquab2024dinov2}, RADIOv2.5~\citep{heinrich2025radiov2} and OpenCLIP~\citep{cherti2023reproducible, nezhurina2025scaling}. \methodname exemplifies a fully \emph{open-source} approach, providing complete transparency from model weights to data and processing methods. {\color{blue}{\boldsymbol{$\sim$}}}: partially; \cellcolor{yellow!20}\umarkc: not specified. NSFW and CSAM are acronyms for "Not Safe For Work" and "Child Sexual Abuse Material," respectively.}
\label{tab:open-source}
\end{center}
\vspace{-16pt}
\end{table*}

Despite the growing importance of VFMs, there is still a lack of \emph{fully open, high-performing, and practical frameworks}. Current state-of-the-art models, including DINOv2~\citep{oquab2024dinov2}, SEER~\citep{goyal2021self}, billion-scale MAE~\citep{singh2023effectiveness}, and SigLIP 2~\citep{tschannen2025siglip2}, rely on proprietary datasets and often withhold critical or all training code, creating a significant barrier to reproducibility, accessibility, and scientific progress. To address this gap, we introduce \methodname, a \emph{fully open (data, weights, code) self-supervised VFM} that not only matches but often surpasses the performance of these proprietary counterparts. A key aspect of our contribution is the release of intermediate checkpoints, which provide unique insight into the training trajectory by enabling the community to analyze convergence behavior and study emergent properties. 
Inspired by Web-SSL's openness, \methodname provides a more accessible framework for models of different scales, intermediate checkpoints and data, while also achieving superior performance (see ~\autoref{sec:results}). \emph{By integrating full openness, high performance, and practical accessibility, \methodname establishes a new standard for transparent VFM research} as shown in~\autoref{tab:open-source}. \alert{For a comprehensive overview of prior VFMs, we refer the readers to ~\autoref{sec:related_work}.}

The strong performance of \methodname stems from two key technical innovations that overcome fundamental limitations in SSL. The first addresses a core shortcoming of models such as DINOv2, which depend on optimal-transport clustering (i.e., Sinkhorn-Knopp) for pseudo-label assignment. 
This process is inherently ambiguous—for instance, vehicles can be grouped by manufacturer, color, or model year—and current methods address this by using very large codebooks (e.g., 131K in DINOv2). To address this, we introduce a multi-head clustering projector using nested Matryoshka representations \cite{kusupati2022matryoshka}, where progressive neuron subsets cluster data into increasingly finer-grained groupings. \alert{This approach not only reduces parameters compared to conventional approaches~\cite{ruan2022weighted, caron2020unsupervised} but also improves performance and decreases memory requirements for downstream tasks} such as k-nearest neighbors classification, leading to higher performances at equal memory.

Second, we address a subtle issue with dense clustering: representations can be biased by patch position rather than semantic content. We introduce a lightweight post-pretraining technique that first learns linear projections to predict patch positions, then projects the latent space to an orthogonal subspace devoid of this positional information. The result is a dense representation space that emphasizes semantics over spatial positioning, leading to substantial gains on challenging benchmarks like dense in-context learning~\citep{balazevic2023towards} and unsupervised semantic segmentation~\citep{van2021unsupervised}.
In summary, our key contributions are:  
\begin{itemize}[leftmargin=20pt, topsep=0pt, parsep=0pt] 
\item We present \methodname, the first fully open-source (code, data, weights) and high-performance VFM that often outperforms proprietary models while ensuring clear accessibility %
\item We introduce two technical innovations: a Matryoshka multi-head clustering approach for high-quality representations, and a spatial-semantic disentanglement post-pretraining technique that refines representations for stronger backbones. Together, these methods significantly advance the base SSL approach, as shown in our analysis (see 
\autoref{tab:rebut-in21k}).  
\item We demonstrate strong performance across diverse tasks: including in-context learning, linear segmentation and object discovery, surpassing DINOv2-G by up to 3\%, outperforming it on OOD detection and 3D understanding, while matching its performance on classification—all without proprietary data.  
\item \alert{We show that~\methodname learns high-quality representations at modest model sizes \emph{without} any teacher supervision, eliminating the need for distillation from a much larger model. This makes our approach more accessible and scalable.}
\end{itemize}

\vspace{-6pt}
\section{Method}
\label{sec:method}

\begin{figure}[t]
    \centering
    \includegraphics[width=\linewidth]{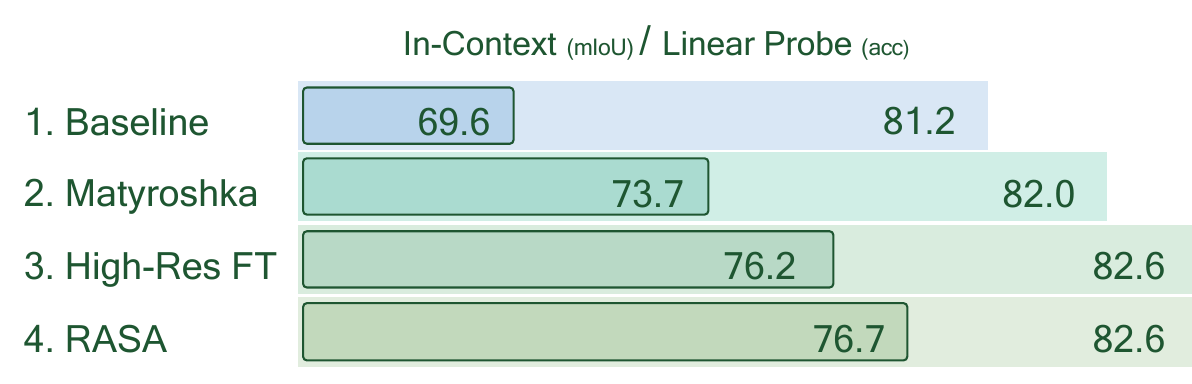}
    \vspace{-10pt}
    \caption{We incrementally add Matryoshka representations, high-resolution fine-tuning and RASA to a DINOv2-B pretrained on ImageNet-21k. Each addition yields consistent performance gains, measured by linear probing on ImageNet-1K (outer bar) and in-context segmentation~\citep{balazevic2023towards} on Pascal VOC (inner bar).} 
    \label{fig:ablation}
    \vspace{-12pt}
\end{figure}

We propose~\methodname, a scalable open-source self-supervised learning (SSL) framework built on DINOv2~\citep{oquab2024dinov2} and pretrained on large public image datasets.
\alert{It tackles key limitations in existing vision SSL models 
through three main components.}
First, \emph{CyclicMask}, inspired by~\citep{darcet2023vision}, is
a masking strategy 
that circularly shifts masked patches to break simple spatial continuity and promote semantic feature learning.
Second, we introduce \emph{Matryoshka Embeddings}~\citep{kusupati2022matryoshka}, a nested multi-head clustering approach that shares projection layers to generate compressed multi-resolution representations; and finally \emph{RASA}, a lightweight post-pretraining step that %
removes feature components correlated with absolute patch positions, resulting in  spatially invariant representations. 
\autoref{fig:ablation} shows that each component provides consistent gains in both in-context segmentation and linear classification. %
These cumulative gains enable \methodname to outperform DINOv2 across a suite of tasks under an equal IN-21K training setup (see  \autoref{tab:rebut-in21k}).

\paragraph{Preliminaries}
We adopt the multi-crop training strategy from DINO~\citep{caron2021emerging}.
An input image is transformed into multiple augmented views (global and local crops). Each view $x$ is then split into $n$ non-overlapping patches, which are embedded into $\mathbb{R}^d$, and a classification token (\texttt{[CLS]}) \(\in \mathbb{R}^d \) is prepended to form the input sequence.
A Vision Transformer (ViT) backbone~\citep{dosovitskiy2020image} processes this sequence, producing \( n+1 \) embeddings (\( n \) patch embeddings and one \texttt{[CLS]} embedding). The same ViT architecture is shared between the student \( f_\theta \) and teacher \( \bar{f}_{\bar{\theta}} \), producing
$Z_s = f_\theta(x) \in \mathbb{R}^{(n+1) \times d}, \quad Z_t = \bar{f}_{\bar{\theta}}(x) \in \mathbb{R}^{(n+1) \times d}$, where \( Z_s \) represents the student's output embeddings and \( Z_t \) represents the teacher's %
embeddings.  
The teacher’s parameters \( \bar{\theta} \) are updated via exponential moving average (EMA) of the student’s parameters.

For supervision, we apply projection heads to the student embeddings \( Z_s \). The \texttt{[CLS]} embedding is passed through a DINO-style head (a 3-layer MLP with softmax over prototypes) that produces image-level prototype scores, while the patch embeddings are processed by an iBOT-style head that produces patch-level prototype scores. For brevity, we denote both heads as \( h_\theta \) for the student and \( \bar{h}_{\bar{\theta}} \) for the teacher (same architecture, EMA-updated). The teacher’s projected outputs are clustered using Sinkhorn-Knopp~\citep{cuturi2013sinkhorn} to produce balanced target distributions. The student is trained to match these targets via cross-entropy loss, denoted as \( \mathcal{L} \).

\subsection{Matryoshka Representations for Efficient Multi-Granular Learning}

\begin{figure*}[t]
    \centering
    \includegraphics[width=0.8\linewidth]{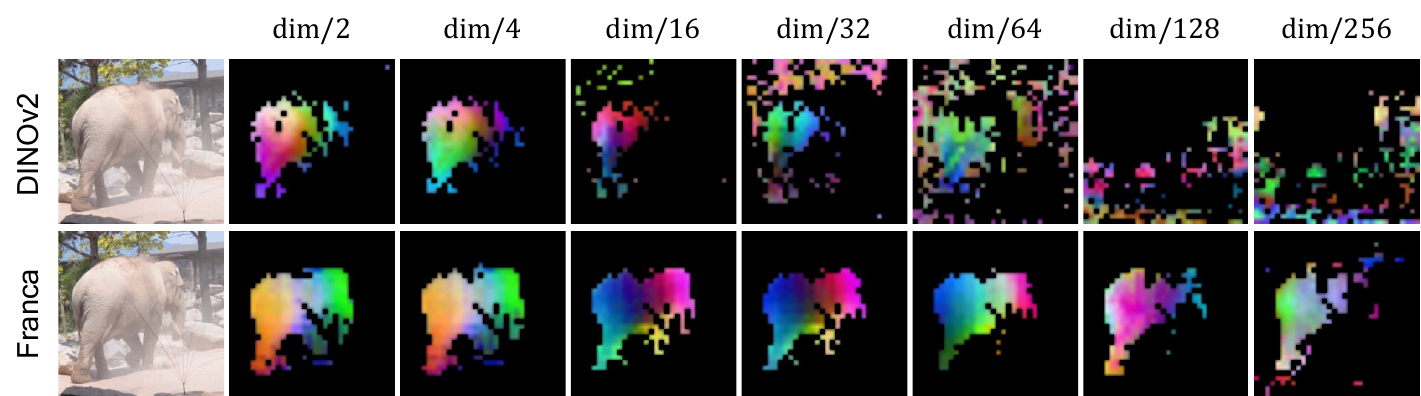}
    \caption{\emph{PCA visualizations across Matryoshka slices}. 
    First three PCA components for different feature slices $m_j$ of \methodname and DINOv2. \methodname being trained only up to $\text{dim}/16$, maintains coherent part structure in smaller feature dimension.}
    \label{fig:per-slice-pca}
    \vspace{-10pt}
\end{figure*}

Standard self-supervised models produce fixed-size embeddings, limiting flexibility under different compute or downstream constraints. To enable adaptable representations across feature granularities, we adopt Matryoshka representations~\citep{kusupati2022matryoshka}, which nest progressively truncated subspaces of a high-dimensional embedding.

Formally, let \( Z_s = f_\theta(x) \in \mathbb{R}^{(n+1) \times d} \) be the unmodified ViT’s output (patch + \texttt{[CLS]} embeddings). We define nested dimensions \( \mathcal{M} = \{m_1, \dots, m_k\} \), where \( m_1 < \cdots < m_k = d \), and extract sub-embeddings $Z_s^{(j)} = Z_s[:, 1:m_j], \quad \forall m_j \in \mathcal{M}.$
  
Each \( Z_s^{(j)} \) is processed by an independent projection head \( h_\nu^{(j)} \), with proportionally fewer prototypes as \( m_j \) decreases. A cross-entropy loss \( \mathcal{L}^{(j)} \) is applied to each head’s output. The total loss is the sum across all levels with equal weights:
$\mathcal{L}_{\text{total}} = \sum_{j=1}^k \mathcal{L}^{(j)}.$ The standard Matryoshka approach~\citep{kusupati2022matryoshka} slices the whole \textit{encoder}’s output along the feature dimension and applies the \emph{same} projection head to each sub-embedding. In contrast, we keep the backbone unmodified and extend this setup by attaching a \emph{dedicated projection head and clustering objective} to each subspace. This allows each slice to produce distinct prototypes and prototype assignments, encouraging specialization across representational granularity of the features across training steps. Our framework supports hierarchical learning: coarse heads capture global semantics, while fine heads focus on local structure akin to early clustering works~\citep{ji2019invariant,asano2020self,vangansbeke2020scan} and unlike most recent representation learning works that optimize only a single feature space~\citep{oquab2024dinov2, tschannen2025siglip2, radford2021learning}.

\begin{figure}
    \centering
\includegraphics[width=0.9\linewidth]{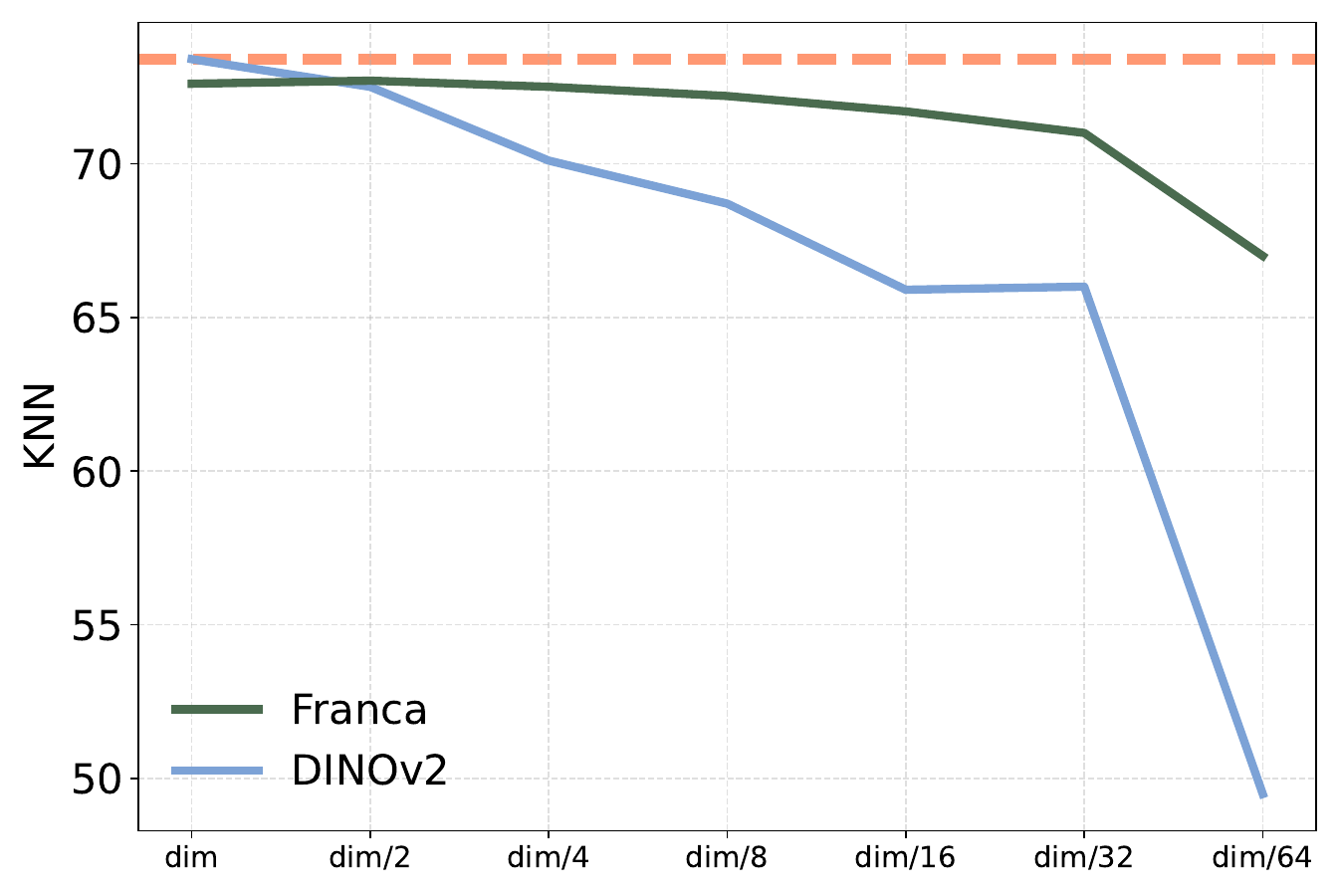}
    \caption{\alert{\emph{k-NN classification accuracy on ImageNet-v2} at varying embedding slice levels using a ViT-L backbone. \methodname consistently outperforms DINOv2 across all subspace dimensions, maintaining high performance even under strong compression ($\text{dim}/64$). DINOv2 was not trained with sliced dimensions and its features are uniformly distributed across the  embedding space.}}
    \label{fig:maty_knn}
\end{figure}
\vspace{-1pt}

As shown in~\autoref{fig:ablation}, our Matryoshka framework yields the largest gains on dense prediction tasks and the PCA visualizations in~\autoref{fig:per-slice-pca} show that ~\methodname preserves coherent part-level structure beyond trained dimensions, whereas DINOv2 loses semantic alignment.  \alert{\autoref{fig:maty_knn} 
shows \methodname significantly outperforms DINOv2~\citep{oquab2024dinov2} across all embedding sizes, especially under heavy compression. The smooth degradation across embedding slices qualitatively supports~\methodname’s hierarchical learning behavior, indicating that higher-level semantic structure is preserved even in lower-dimensional subspaces. For fairness, we note that DINOv2 was not trained for dimensional truncation; its information is spread uniformly across the feature space.}

\subsection{Balancing Spatial Distribution of Visible Patches with CyclicMask}
\begin{figure}[t]
    \centering
    \includegraphics[width=\linewidth]{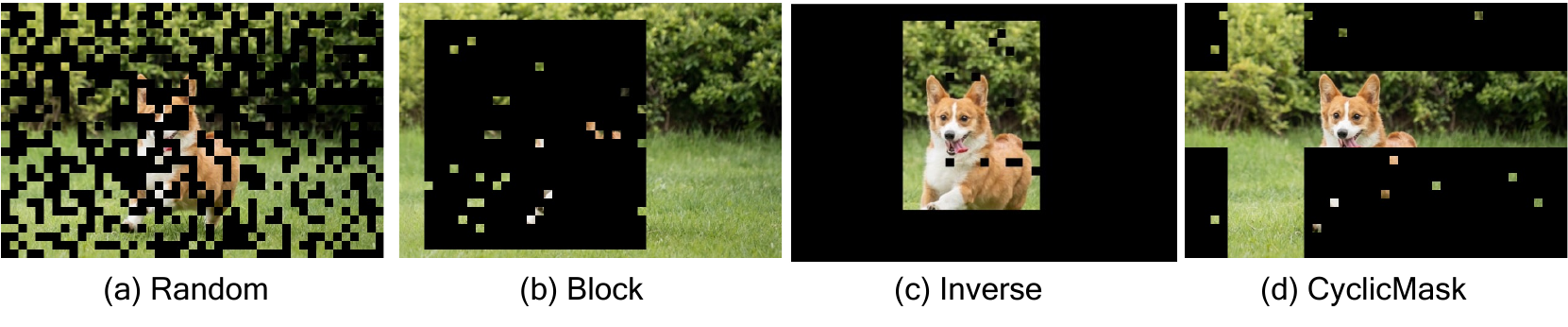}
    \vspace{-10pt}
    \caption{\emph{Masking strategies used in masked image modeling.} Compared to Random (a), Block (b), and Inverse (c) masking, our  CyclicMask (d) circularly shifts the visible region across spatial axes,  preventing the model from being biased toward specific spatial locations.} 
    \label{fig:masking-strat}
    \vspace{-12pt}
\end{figure}

Masked image modeling (MIM) is a core component in many self-supervised vision frameworks~\citep{oquab2024dinov2, zhou2021ibot, zhou2022mugs}, where a portion of input patches are masked and the model learns to predict the unmasked regions. Commonly adapted strategies include \emph{random masking} and \emph{block masking}, where patches are masked randomly or as a block across the image. While simple and stochastic, this approach lacks spatial structure, often leading to fragmented visible regions that provide limited contextual coherence as shown in~\autoref{fig:masking-strat} (a) and (b).

\subsection{RASA: Removal of Absolute Spatial Attributes}

\begin{figure*}[t]
  \centering
  \includegraphics[width=0.85\textwidth]{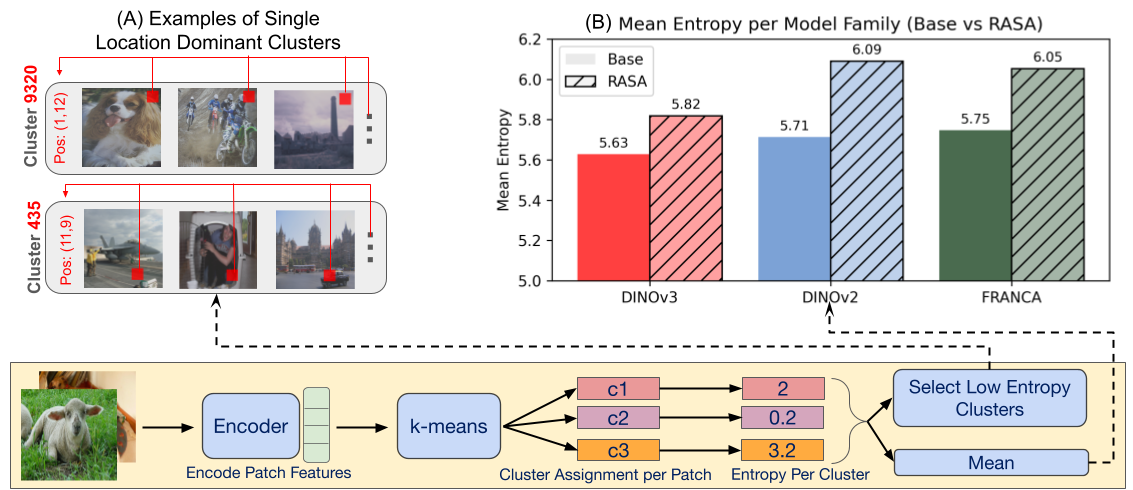}
  
  \vspace{-5pt}
  \caption{\emph{Entropy of patch locations per cluster.} For each visual cluster out of the $65k$ clusters from $k$-means, we compute the entropy of the 2D coordinates of its assigned patches (bottom row). Lower Mean Entropy (\textsc{B}: top right) indicates positional bias in its representations. We find that DINOv2 has many low-entropy clusters firing at fixed locations (\textsc{A}: top left). 
  Top right: Our RASA post-training  increases  spatial entropy, thereby debiasing positional information from patch features. 
  }
  \label{fig:patch_entropy}
\end{figure*}

ViT models often develop unintended spatial biases from their fixed patch layouts and positional embeddings, entangling location with semantic content. Our preliminary study in \autoref{fig:patch_entropy} demonstrates this issue: using a frozen model on COCO images, we assigned each visual patch to $65k$ clusters from $k$-means and computed the spatial entropy of patch locations per cluster (\autoref{fig:patch_entropy}: bottom). 
DINOv2 patch clusters are frequently centered at fixed positions (\autoref{fig:patch_entropy}-A: top-left), exhibiting low mean spatial entropy (\autoref{fig:patch_entropy}-B: top-right), suggesting that cluster assignments are driven by location than semantics. Our RASA post-training increases spatial entropy (\autoref{fig:patch_entropy}-B: top-right), resulting in less location-driven clusters.

\tdplotsetmaincoords{60}{120}
\definecolor{myred}{RGB}{232,168,161}
\definecolor{myblue}{RGB}{124,162,214}
\definecolor{mygreen}{RGB}{74,107,79}
\definecolor{myviolet}{RGB}{65,41,145}
\definecolor{myyellow}{RGB}{255,192,0}

\noindent
\begin{figure}
\centering
\begin{tikzpicture}[tdplot_main_coords, scale=3]

\draw[->] (0,0,0) -- (1.2,0,0) node[anchor=north east]{$x$};
\draw[->] (0,0,0) -- (0,1.2,0) node[anchor=north west]{$y$};
\draw[->] (0,0,0) -- (0,0.0,0.9) node[anchor=south]{$z$};

\coordinate (Z0) at (1.1,1.3,1.4);
\coordinate (P1) at (0.6,0.7,0.1);
\coordinate (Z1) at (0.4,0.7,0.3);
\coordinate (ur) at (1.2,0.35,0);
\coordinate (uc) at (0.2,1.0,0.2);

\coordinate (corner1) at (ur);
\coordinate (corner2) at (uc);
\coordinate (corner3) at ($(ur)+(uc)$);
\coordinate (corner0) at (0,0,0);

\filldraw[fill=gray!20, opacity=0.3, draw=gray!40] 
    (corner0) -- (corner1) -- (corner3) -- (corner2) -- cycle;
\node[gray!60] at (1.1,1.6,0.2) {\scriptsize Positional Plane};

\draw[->, thick, myblue!110] (0,0,0) -- (Z0) node[anchor=south east]{$Z_i^{(0)}$};
\draw[->, thick, myred!140] (0,0,0) -- (ur) node[anchor=south east]{$u_r$};
\draw[->, thick, myyellow] (0,0,0) -- (uc) node[anchor=south west]{$u_c$};
\draw[->, thick, mygreen!110, dashed] (0,0,0) -- (P1) node[anchor=north west,xshift=-15pt,yshift=0pt]{$p_i$};
\draw[->, thick, myviolet!90, dashed] (P1) -- (Z0) node[anchor=west,xshift=-5pt,yshift=-15pt]{$Z_i^{(1)}$};

\end{tikzpicture}
\caption{Each iteration of RASA projects a patch embedding $Z_i$ onto a learned positional plane $\text{span}\{u_r, u_c\}$ and subtracts its projection $p_i$.}
\label{fig:rasa-gram-schmidt-plane2}
\vspace{-12pt}
\end{figure}

To directly address the spatial entanglement, we propose Removal of Absolute Spatial Attributes (\emph{RASA}), a post-training method designed to disentangle spatial information from patch embeddings. After pretraining, we process the patch features $Z$ (for brevity, we denote $Z_t$ as $Z$ throughout this section) through an alternating optimization procedure. At iteration $t$, the input consists of the $n$ patch embeddings $Z^{(t)} = \{ Z_i \in \mathbb{R}^D \}_{i=1}^{n}$. 
We first optimize a linear position prediction head, parametrized by a matrix $W \in \mathbb{R}^{2 \times D}$, on a small set of images to predict normalized patch coordinates via a sigmoid $\sigma(\cdot)$, minimizing the mean squared error: 
$\mathcal{L}_{\text{pos}} = \tfrac{1}{n} \sum_{i=1}^n \lVert \sigma(W Z_i) - y_i \rVert_2^2,$
where $y_i \in [0, 1]^2$ are the normalized 2D coordinates of patch $i$. 
The row vectors of $W$ are orthonormalized using Gram–Schmidt \citep{golub2013matrix} to form the basis vectors $u_r$ and $u_c$, which span the positional subspace (visualized in gray in \autoref{fig:rasa-gram-schmidt-plane2}).
We then remove the component of each feature vector that lies within this positional subspace. This is achieved by projecting $Z_i$ onto the subspace and subtracting that projection, yielding a refined embedding $Z^{(t+1)}_i$ that is less aligned with positional information while retaining semantic content:  
\begin{align}
& p_i = \langle Z_i, u_r \rangle u_r + \langle Z_i, u_c \rangle u_c, \\
&Z^{(t+1)}_i = Z^{(t)}_i - p^{(t)}_i. \label{eq:rasa}
\end{align}
This iterative refinement process, summarized in \autoref{fig:rasa-gram-schmidt-plane2}, continues until $\mathcal{L}_{\text{pos}}$ saturates (typically within $9$ iterations), effectively removing linearly predictable spatial bias while preserving semantic content.
Due to its linear construction, the entire RASA transformation for a single iteration $t$ can be expressed as multiplication by a single matrix $L^{(t)}$:
\begin{align}
     Z^{(t+1)}_i = Z^{(t)}_i L^{(t)} = Z^{(t)}_i (I- p^{(t)}_i)  = Z^{(t)}_i(I - u_r u_r^\top + u_c u_c^\top),
\end{align}
The complete multi-step transformation is the product of these matrices, $L = \prod_{t=1}^{T} L^{(t)}$. This final matrix $L$ can be appended to the weights of the final ViT layer, resulting in no architectural changes and zero inference overhead.

\section{Experimental Results}
\label{sec:results}
\vspace{-3pt}

\paragraph{Training setup}
We train~\methodname using ViT-B, ViT-L, and ViT-G encoders with patch size 14 and without registers. We pretrain from scratch for 625K iterations with batch size 2048 for ViT-B and 3072 for ViT-L and ViT-G using an image resolution $224\times224$ with \emph{Matryoshka Embeddings} and \emph{CyclicMask} inspired from~\cite{darcet2025capi}. ViT-L and ViT-G is pretrained on LAION-600M, while ViT-B (for ablations) uses ImageNet-21K. We then finetune the models with batch size 1024 at $364\times364$ and $518\times518$ resolutions (for 30K and 10K iterations, respectively) on a mix of ImageNet-1K, ADE20K, COCO, KITTI, and VOC; due to computational constraints, ViT-G skips this stage. Finally, we apply the lightweight \emph{RASA} post-training on Pascal VOC, for $8$ iterations, using a learning rate of $0.002$, and a batch size of $128$. More details are in \autoref{sec:app-setup}.

\begin{table}[ht!]
\centering
\footnotesize
\setlength{\tabcolsep}{2pt}
\begin{tabular}{lcccccc}
\toprule
\textsc{Method} & \textsc{Arch.}    & \textsc{HRFT}       & \textsc{Knn} & \textsc{IN-val} & \textsc{In-Context} & \textsc{VOS} \\
                &                   & \textsc{+RASA}      & (IN-1K)      & (IN-1K)         &  (ADE20K)           & (DAVIS)      \\
\midrule
DINOv2          & ViT-B/14          & \ding{55}            & 77.0       & 81.2            & 30.0              & 63.1                      \\
\rowcolor{rowmint}
Franca          & ViT-B/14          & \ding{55}            & \textbf{77.5}       & \textbf{82.0}            & \textbf{31.6}                & \textbf{65.5}                      \\
\cmidrule{2-7}
DINOv2          & ViT-L/14          & \ding{55}            & 82.1       & 84.0            & 33.5                 & 64.2                      \\
\rowcolor{rowmint}
Franca          & ViT-L/14          & \ding{55}            & \textbf{82.2}       & \textbf{84.5}            & \textbf{34.5}                 & \textbf{68.0 }                     \\
\midrule
DINOv2          & ViT-B/14          & \checkmark            & 76.8       & 81.0            & 32.4                 & 69.2                    \\
\rowcolor{rowmint}
Franca          & ViT-B/14          & \checkmark            & \textbf{79.6}       & \textbf{82.6}            & \textbf{35.0}              & \textbf{70.6}                    \\
\cmidrule{2-7}
DINOv2          & ViT-L/14          & \checkmark            & 80.7       & 84.5            & 37.9                 & 66.6                    \\
\rowcolor{rowmint}
Franca          & ViT-L/14          & \checkmark            & \textbf{82.5}       & \textbf{85.2}            & \textbf{39.6}             & \textbf{70.0}                      \\
\bottomrule
\end{tabular}
\caption{\alert{
\emph{Controlled comparison with DINOv2 on IN-21K.}  We compare~\methodname and DINOv2 using ViT-B and ViT-L backbones, pretrained on the same IN-21K dataset. All models are trained with identical hyperparameters and without distillation. Across classification and dense prediction tasks,~\methodname consistently outperforms DINOv2, often by a significant margin.}}
\label{tab:rebut-in21k}
\end{table}

\subsection{Controlled Comparison with DINOv2}
\alert{
We begin by conducting a controlled comparison between \methodname and DINOv2 to study the effect of architectural differences. We use ViT-B/14 and ViT-L/14 backbones pretrained on IN-21K with identical hyperparameters—and crucially, without high-resolution fine-tuning or distillation from a ViT-G model—we isolate the impact of \methodname's architectural and training innovations. As shown in \autoref{tab:rebut-in21k}, 
\methodname consistently outperforms DINOv2 across a suite of tasks, in particular dense segmentation (ADE20K, DAVIS), demonstrating that the Matryoshka, CyclicMask, and RASA components lead to better semantically aligned ViT representations that are more effective for dense tasks.
}

\subsection{Dense tasks}
\paragraph{In-Context Learning}

\begin{table*}[t]
\centering
\scriptsize
\setlength{\tabcolsep}{5pt}
\begin{subtable}[t]{0.5\textwidth}
\centering
\begin{tabular}{llcc|cc}
\toprule
\multirow{2}{*}{\textsc{Method}} & \multirow{1}{*}{\textsc{Backbone}} & \multicolumn{2}{c}{\textsc{Lin. Seg.}} & \multicolumn{2}{c}{\textsc{In-context}} \\
\cmidrule{2-6}
 &                             & \textsc{VOC}  & \textsc{ADE20K} & \textsc{VOC}  & \textsc{ADE20K} \\
\midrule
SigLIP            & ViT-B/16   & 57.8          & 23.1            & 33.9          & 10.6 \\
SigLIP 2           & ViT-B/16   & 69.6          & 23.1            & 65.0            & 32.3     \\
iBOT              & ViT-B/16   & 73.1          & 30.1            & 66.6          & 26.9 \\
EVA-CLIP          & ViT-B/16   & 70.4          & 34.6            & 34.8          & 11.3 \\
DINOv2$^\dagger$  & ViT-B/14   & 86.9          & 41.3                & 69.6          & 30.0 \\
DINOv2$^\S$       & ViT-B/14   & \textbf{90.2} & \textbf{49.7}    & \textbf{77.1} & \textbf{37.7} \\
\cmidrule{2-6}
\rowcolor{rowmint}
~\methodname (ours) & ViT-B/14 & \underline{89.4}          & \underline{46.2}            & \underline{76.7}          & \underline{35.0} \\
\midrule
SigLIP 2          & ViT-L/16   & 66.6          & 29.8            & 46.4          & 21.3 \\
Web-SSL           & ViT-L/14   & \textbf{92.3}          & 46.3            & 71.3          & 35.3 \\
DINOv2$^\dagger$  & ViT-L/14   & 89.3            & 45.4              & 72.0          & 33.5 \\
DINOv2$^\S$       & ViT-L/14   & 90.3            & \textbf{50.7}   & \underline{74.6}          & \underline{38.6} \\
\cmidrule{2-6}
\rowcolor{rowmint}
~\methodname (ours) & ViT-L/14 & \underline{90.5 }         & \underline{48.9}            & \textbf{79.5} & \textbf{39.6} \\
\midrule
Web-SSL           & ViT-G/14   & 89.5          & \textbf{46.7}            & 73.3          & \underline{36.7} \\
DINOv2            & ViT-G/14   & \textbf{90.6}            & 46.2            & \underline{73.7}          & \textbf{37.7} \\
\cmidrule{2-6}
\rowcolor{rowmint}
~\methodname (ours) & ViT-G/14 & \underline{90.2} & \underline{46.5}   & \textbf{76.7} & 36.5 \\
\bottomrule
\end{tabular}
\caption{Linear and In-context Segmentation.}
\label{tab:segmentation}
\end{subtable}
\hfill
\begin{subtable}[t]{0.45\textwidth}
\centering
\begin{tabular}{lcc|cc}
\toprule
                &                & \textsc{Video Obj. Segm.}        & \multicolumn{2}{c}{\textsc{TokenCut}} \\
\cmidrule{2-5}
\textsc{Method} & \textsc{Arch.} & $\mathcal{(J+F)}_m$ & VOC07  & VOC12 \\
\midrule
RADIOv2.5       & ViT-B/16       & 66.7                    &    44.6    &   48.1   \\
C-RADIOv3       & ViT-B/16       & 67.7                    &   47.3    &  50.8    \\ 
DINOv2$^\S$          & ViT-B/14       & 68.5                    & 42.9      & 48.3 \\
DINOv2-R$^\S$        & ViT-B/14       & 67.8                      & \underline{53.0}    & \underline{56.9} \\
\cmidrule{2-5}
\rowcolor{rowmint}
\methodname          & ViT-B/14       & \textbf{70.6}                     & \textbf{53.2}   & \textbf{59.1} \\
\midrule
RADIOv2.5       & ViT-L/16       & 67.5                    &    51.3    &  54.6    \\
C-RADIOv3       & ViT-L/16       & 67.3                     &  --   &  --   \\
DINOv2$^\S$      & ViT-L/14      & \textbf{69.1} & 41.3 &  47.2 \\
DINOv2-R$^\S$        & ViT-L/14       & 66.9 & \underline{57.6} & \textbf{61.3} \\
\cmidrule{2-5}
\rowcolor{rowmint}
\methodname          & ViT-L/14       & \textbf{69.1} & \textbf{59.5} & \underline{61.0} \\
\bottomrule
\end{tabular}
\caption{Unsupervised Video Object Segmentation on DAVIS and  Object Discovery using TokenCuT.}
\label{tab:vos_tokencut}
\end{subtable}
\vspace{-6pt}
\caption{\emph{Segmentation Benchmarks.} Left: Linear and In-Context Segmentation. Right: Video object segmentation on DAVIS2017 and Unsupervised Object Discovery using TokenCuT (VOC 07/12).  
$^\dagger$: reproduced on IN-21K, without distillation; $\S$ : distilled from DINOv2(-R)-G on LVD-142M. from LVD-1.7B \textbf{bold}: best; \underline{underline}: second best
}
\label{tab:combined_all}
\end{table*}

We evaluate~\methodname on the Hummingbird Benchmark~\citep{balazevic2023towards, pariza2024hbird}, 
which frames semantic segmentation as a nearest-neighbor retrieval task using patch features, without model fine-tuning.
We report mean Intersection-over-Union (mIoU) on Pascal VOC~\citep{pascal-voc-2012} and ADE20K~\citep{zhou2017scene} in \autoref{tab:segmentation}.

\methodname consistently achieves strong segmentation results. For instance, with ViT-L/14, it surpasses DINOv2-L (distilled from DINOv2-G and trained on LVD-142M) and Web-SSL by up to +5\% and +8\% mIoU on Pascal VOC, and by +1\% and +4\% on ADE20K, respectively. 
\alert{Notably, DINOv2's pretraining data (LVD-142M) includes VOC and ADE20K, whereas~\methodname only uses them during the limited high-resolution finetuning stage (which \methodname's ViT-G skips). Despite no exposure to these datasets,~\methodname-G achieves comparable performance or outperforms DINOv2. The performance gains also increases with model capacity from ViT-B to ViT-L. }
These results demonstrate \methodname's ability to learn spatially precise, semantically meaningful features that transfer effectively to segmentation without fine-tuning. We also report overclustering performance, which measures semantic alignment of spatial features in~\autoref{tab:overclustering_results}.

\begin{figure}[h]
    \centering
    \includegraphics[width=\linewidth]{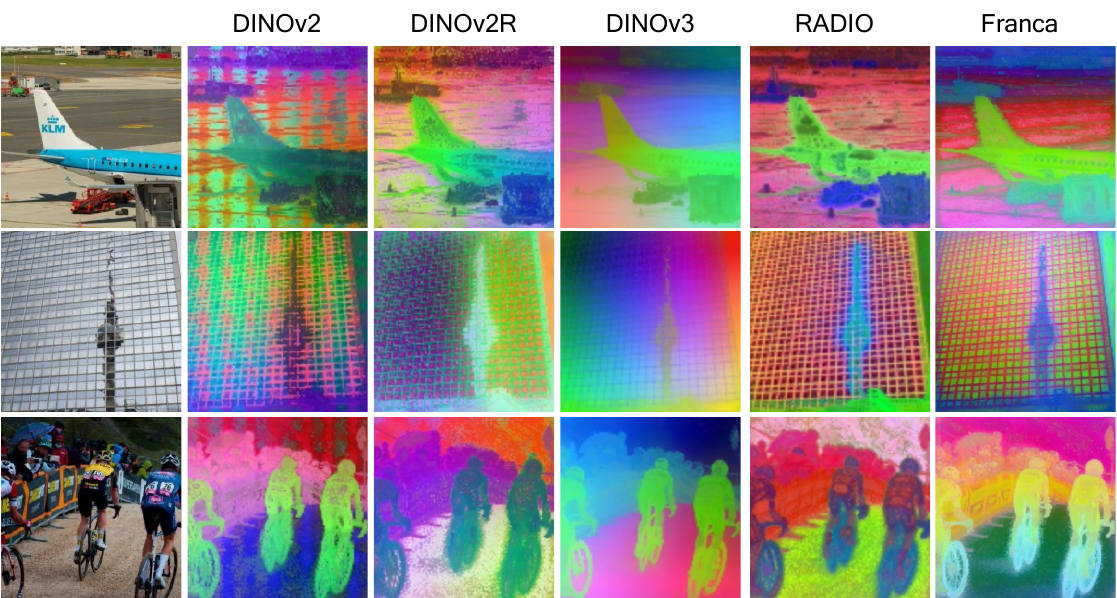}
    \vspace{-12pt}
    \caption{
    \emph{Semantic Consistency of Dense Features.} Visualization of dense features via PCA (first three components as RGB) demonstrates that our \methodname features consistently match semantic parts with the same color, despite variations in pose, style, or even object identity (e.g., the consistent coloring for the bikers and bikes in the last row). Images were randomly sampled from the Segment Anything-1B dataset~\citep{kirillov2023segment} (\texttt{np.random.randint(seed=42)}).
    }
    \label{fig:pca_viz}
    \vspace{-12pt}
\end{figure}

\paragraph{Linear segmentation}
\autoref{tab:segmentation} also evaluates \methodname on semantic segmentation under a linear probing protocol. \methodname achieves strong results across all backbones. With ViT-B/14, it performs on par with DINOv2-B, which is distilled from DINOv2-G and pretrained on LVD-142M. Crucially, a non-distilled DINOv2-B model trained on ImageNet-21K (the same setting as \methodname for ViT-B) performs substantially worse (86.9 vs. 89.4 mIoU on VOC; 41.3 vs. 46.2 on ADE20K). This suggests the strong performance of the official DINOv2-B is largely due to distillation. With ViT-G/14, \methodname matches DINOv2 and outperforms Web-SSL. Notably, \methodname surpasses Web-SSL on several benchmarks despite Web-SSL's pretraining on a substantially larger dataset (2B images), highlighting our approach's effectiveness. Interestingly, SigLIP 2~\citep{tschannen2025siglip2} performs poorly on segmentation, particularly on VOC (57.8 mIoU), suggesting limited spatial localization capabilities despite its long training and strong image-text alignment.

\paragraph{Video Object Segmentation}
We extend our evaluation to video object segmentation (VOS) on DAVIS~\citep{pont2017}, where the goal is to propagate a ground-truth mask from the first frame through a video using feature similarity. From~\autoref{tab:vos_tokencut}, \methodname achieves the best results across all backbones. With ViT-B, it scores 70.6\%, surpassing 
DINOv2 by 2\%.
With ViT-L, \methodname matches or outperforms other methods, reaching 69.1\%. This shows that \methodname produces temporally stable features that enable consistent tracking.

\begin{figure*}[h!]
    \centering
    \includegraphics[width=0.9\linewidth]{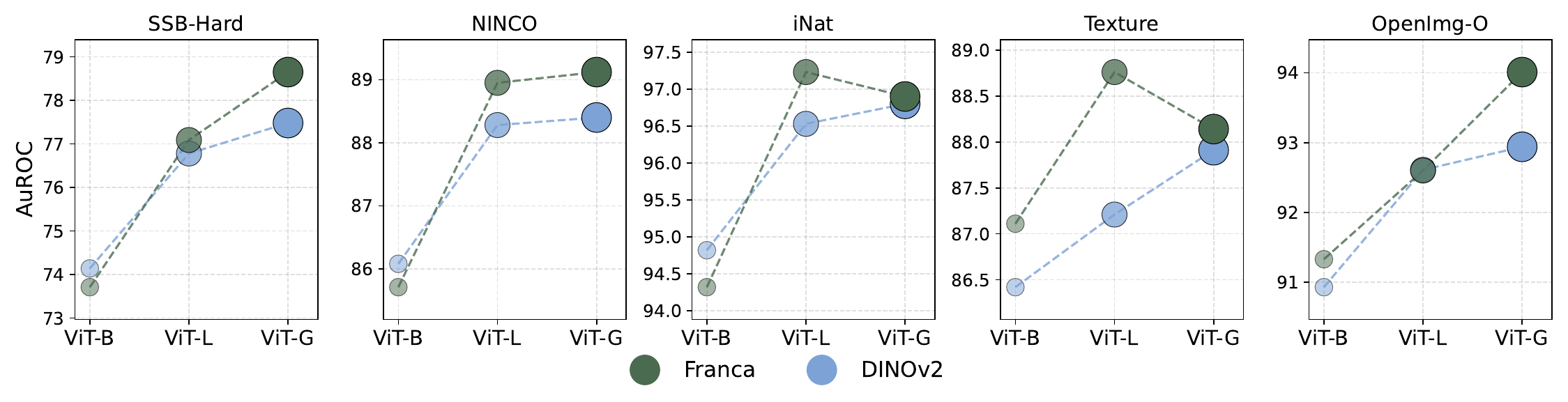}
    \vspace{-6pt}
    \caption{\emph{Out-of-Distribution Detection} across five robustness-benchmarks: SSB-Hard~\citep{vaze2021open}, NINCO~\citep{bitterwolf2023ninco}, iNaturalist~\citep{huang2021mos}, OpenImage-O~\citep{wang2022vim}, and Texture~\citep{kylberg2011kylberg}. ~\methodname consistently outperforms DINOv2, at larger scales, demonstrating its robustness across distribution shifts. DINOv2-B and DINOv2-L are distilled from DINOv2-G and trained on LVD 142M, while neither variants of~\methodname are distilled. 
    } 
    \label{fig:ood_plots}
\end{figure*}

\begin{table*}[t]
\centering
\footnotesize
\begin{tabular}{llllccc|ccc}
\toprule
\multicolumn{7}{c}{\textsc{Classification}} & \multicolumn{3}{c}{\textsc{Robustness}} \\
\cmidrule(r){1-7} \cmidrule(l){8-10}
\textsc{Method} & \textsc{Arch.} & \textsc{Data} & \textsc{Text} & \textsc{Knn} & \textsc{IN-val} & \textsc{v2} & IN-A & IN-R & Sketch \\
\midrule
IBoT               & ViT-B/16     & IN-21K       & \xmarkc & 77.1 & 79.5   & --     & --   & --   & --   \\
DINOv2$^\dagger$   & ViT-B/14     & IN-21K       & \xmarkc & 77.0 & 81.2   & 70.9       & 44.1 & 50.1  & 40.8  \\
DINOv2$^\S$        & ViT-B/14     & LVD-142M     & \xmarkc & \textbf{82.1} & \textbf{84.5} & \textbf{75.1} & \textbf{55.1} & \textbf{63.3} & \textbf{50.6} \\
\cmidrule(r){2-10}
\rowcolor{rowmint}
\methodname (ours) & ViT-B/14     & IN-21K       & \xmarkc & \underline{79.5} & \underline{82.6} & \underline{73.7} & \underline{48.5} & \underline{54.6} & \underline{44.1} \\
\midrule 
SigLIP             & ViT-L/16     & WebLI        & \cmarkc & --  & 80.5 & 74.2 & 76.5 & 95.0 & \underline{73.6} \\
SigLIP 2            & ViT-L/16     & WebLI        & \cmarkc & --  & 82.5 & 76.8 & \underline{84.3} & \textbf{95.7} & \textbf{75.5} \\
PE$_{\text{core}}$ & ViT-L/16     & MC-2B        & \cmarkc & \textbf{83.5} & 83.9 & \underline{77.9} & \textbf{89.0} & \underline{95.2} & 73.4 \\
Web-SSL            & ViT-L/16     & MC-2B        & \xmarkc & 76.8 & 82.4 & 71.0 & 67.3 & 68.9 & 54.8 \\
DINOv2$^\dagger$   & ViT-L/14     & IN-21K       & \xmarkc & \underline{82.1} & 84.0   & 75.5              & 61.5              &  61.0          & 45.4 \\
DINOv2$^\S$        & ViT-L/14     & LVD-142M     & \xmarkc & \textbf{83.5} & \textbf{86.3} & \textbf{78.0} & 71.3 & 74.4 & \textbf{59.3} \\
\cmidrule(r){2-10}
\rowcolor{rowmint}
\methodname (ours) & ViT-L/14     & LAION-600M    & \xmarkc   & 81.6   & 84.9 & 76.0 & 66.5 & 64.3 & 56.6 \\
\midrule
OpenCLIP           & ViT-G/14     & LAION-2B     & \cmarkc & \textbf{83.2} & \textbf{86.2} & \underline{77.2} & 63.8 & \textbf{87.8} & \textbf{66.4} \\
DINOv2             & ViT-G/14     & LVD-142M     & \xmarkc & \underline{83.1} & \underline{86.0} & \textbf{77.9} & \textbf{75.9} & \underline{78.8} & \underline{62.5} \\
Web-SSL            & ViT-G/14     & MC-2B        & \xmarkc & 79.2 & 84.7 & 74.3 & 73.3 & 75.9 & 60.9 \\
\cmidrule(r){2-10}
\rowcolor{rowmint}
\methodname (ours) & ViT-G/14     & LAION-600M   & \xmarkc & 83.0 & \underline{86.0} & \textbf{77.9} & \underline{75.6} & 75.8 & 60.6 \\
\bottomrule
\end{tabular}
 \vspace{-3pt}
\caption{\emph{Classification and Robustness.} performance across vision-language and vision-only models. We report top-1 linear probing accuracy on IN-1K (val, ReaL, v2) and robustness benchmarks (IN-A, IN-R, Sketch). ~\methodname, trained without text supervision, matches or exceeds the performance of larger text-supervised models and outperforms DINOv2 when reproduced on the same data and training strategy. $\dagger$: reproduced on IN-21K without distillation; $\S$: distilled from DINOv2-G on LVD-142M. \textbf{Bold}: Best; \underline{Underline}: second best.}
 \label{tab:classic_tasks}
\end{table*}

\paragraph{Unsupervised Object Discovery}
We further evaluate unsupervised object discovery using TokenCut~\citep{wang2022tokencut} in~\autoref{tab:vos_tokencut}, which segments objects by leveraging patch-level feature similarity. On VOC07, \methodname-B achieves a higher CorLoc than DINOv2. On VOC12, \methodname obtains 59.1\% CorLoc, outperforming DINOv2 (48.3\%) and DINOv2-R (56.9\%). This demonstrates that \methodname provides spatially coherent features that facilitate unsupervised object segmentation.

\paragraph{PCA of Patch Features}
We visualize patch tokens by projecting them into a 3D RGB color space using PCA. As shown in~\autoref{fig:pca_viz}, the resulting color maps reveal a stark contrast. For DINOv2, PCA highlights only a few scattered high-variance patches (outliers), failing to cover the full object and yielding a fragmented, noisy segmentation. In contrast, \methodname produces dense, coherent color segments aligned with the actual object. Its contours are sharply delineated, and semantically similar parts (e.g., bike, body of aeroplane) receive consistent colors across instances—patterns that do not emerge in DINOv2. Importantly, neither model was trained on the Segment Anything-1B~\citep{kirillov2023segment} subset used for these visualizations, confirming that the segmentation patterns are entirely emergent from the self-supervised features. \alert{This qualitative coherence aligns with the quantitative gains in TokenCut (\autoref{tab:vos_tokencut}), both reflecting \methodname’s spatially consistent patch features.} We visualize the self-attention of different models in~\autoref{fig:attention-viz} (Appendix), showing that \methodname's attention maps provide better object localization.

\begin{table*}[h]
\centering
\footnotesize
\setlength{\tabcolsep}{5pt}
\begin{tabular}{lccc|ccc|ccc}
\toprule
\textsc{Model}  & \multicolumn{3}{c|}{\textsc{Geometry}} & \multicolumn{3}{c|}{\textsc{Texture}} & \multicolumn{3}{c}{\textsc{Geometry+Texture}} \\
\cmidrule{2-10}
 & PSNR $\uparrow$ & SSIM $\uparrow$ & LPIPS $\downarrow$ & PSNR $\uparrow$ & SSIM $\uparrow$ & LPIPS $\downarrow$ & PSNR $\uparrow$ & SSIM $\uparrow$ & LPIPS $\downarrow$ \\
\midrule
EVA-CLIP      & 19.01 & 0.62 & 0.35 & 18.02 & 0.60 & 0.32                   & 18.85 & 0.63 & 0.38 \\
CLIP          & 19.47 & 0.64 & 0.33 & 18.04 & 0.60 & 0.32                   & 19.42 & 0.64 & 0.39 \\
SigLIP 2       & 19.39 & 0.64 & 0.33 & \textbf{18.11} & 0.60 & \textbf{0.31} & 19.36 & 0.64 & 0.41 \\
AIMv2         & 19.16 & 0.62 & 0.35 & 18.05 & 0.60 & 0.32                   & 19.04 & 0.63 & 0.39 \\
WebSSL        & 19.56 & 0.64 & 0.33 & 18.05 & 0.60 & 0.32                   & 19.43 & 0.64 & 0.37 \\
DINOv2        & 19.34 & 0.63 & 0.34 & 17.96 & 0.60 & 0.33                   & 19.31 & 0.64  & \textbf{0.36} \\
\cmidrule{2-10}
\rowcolor{rowmint}
\methodname & \textbf{19.58} & \textbf{0.65} & \textbf{0.32} & 17.97 & \textbf{0.62} & 0.32 & \textbf{19.53} & \textbf{0.65} & 0.37 \\
\bottomrule
\end{tabular}
\vspace{-3pt}
\caption{\emph{Probing with Gaussian Splatting}, Normalized average metrics using Feat2GS~\citep{chen2025feat2gs} across six datasets for two probing schemes: Geometry, Texture and All: Geometry + Texture with ViT-L backbone. We measure PSNR, SSIM (higher is better) and LPIPS (lower is better).~\methodname outperforms SoTA VFMs suggesting strong geometrical awareness.}
\label{tab:feat2gs}
\vspace{-10pt}
\end{table*}

\subsection{Classification and Robustness}

\paragraph{Image Classification}
We evaluate the global image representations learned by \methodname for image classification using nearest-neighbor (KNN) and linear probing protocols on ImageNet-1K. For linear probing, we report results on both the standard validation set and ImageNet-v2~\citep{recht2019imagenet}. As shown in \autoref{tab:classic_tasks}, \methodname achieves performance comparable to DINOv2, even for ViT-B and ViT-L despite the fact that for these variants \methodname does not use distillation from a ViT-G teacher (as DINOv2 does), and the ViT-B model is trained only on the 13M images of IN-21K. \methodname also outperforms Web-SSL~\citep{fan2025scaling}, which uses a much larger dataset (MetaCLIP-2B). Notably, our \methodname-G model matches the linear accuracy of Web-SSL-7B (85.9\% vs. 86.0\%) with 7× fewer parameters.

To ensure a fair comparison, \autoref{tab:classic_tasks} and
\autoref{tab:rebut-in21k} include DINOv2 models (ViT-B/14, ViT-L/14) re-trained on IN-21K without distillation from a ViT-G teacher, using our training setup. These models perform significantly worse than the original DINOv2, indicating that the ViT-G distillation contributes substantially to their performance. In contrast, \methodname achieves strong results without such distillation or extra supervision, matching or surpassing these reproduced DINOv2 versions.

\paragraph{Robustness}
We assess robustness by applying the linear classification heads to the validation sets of ImageNet-A~\citep{hendrycks2021natural}, ImageNet-R~\citep{hendrycks2021many}, and Sketch~\citep{wang2019learning}, which introduce semantic or stylistic variations. From \autoref{tab:classic_tasks},~\methodname demonstrates strong robustness,  \alert{where \methodname-G achieves performance comparable to DINOv2-G across all three datasets} and outperforms OpenCLIP-G by 9\% on ImageNet-A, despite OpenCLIP being trained on over 30× more data and that it's weakly aligned.

\paragraph{Out-of-Distribution Detection}
We further evaluate \methodname on out-of-distribution detection using the OpenOOD benchmark~\citep{zhang2023openood} reporting the AuROC metric across five datasets.~\autoref{fig:ood_plots} shows that \methodname consistently outperforms DINOv2 across large and giant model variants, demonstrating strong robustness to distribution shifts and effective scaling for OOD detection.

\subsection{Probing 3D awareness}
We evaluate texture and geometry awareness using the Feat2GS framework~\citep{chen2025feat2gs}, which leverages novel view synthesis as a proxy for 3D understanding. Features from visual foundation models are mapped into 3D Gaussians through a lightweight readout trained with a photometric loss. For fair comparison, inputs are resized to 512, feature maps are upsampled to 512, and PCA reduces dimensionality to 256. Evaluation uses PSNR, SSIM, and LPIPS under three setups: (a) geometry features predict geometry, with texture optimized; (b) texture features predict texture, with geometry optimized; and (c) all features predict both across six datasets, averaging results over five runs. From ~\autoref{tab:feat2gs}, \methodname achieves the best performance in Geometry and All, indicating strong 3D geometric awareness, while all methods perform similarly on Texture, with SigLIP 2 slightly ahead. We also show the results on dense keypoint matching and depth prediction in~\autoref{tab:3d_understanding}.

\vspace{-3pt}
\section{Conclusion}
In this work, we present \methodname, a new Vision Foundation Model that is open-weight, open-data and open-code. We build this model using a novel Matryoshka-nested clustering self-supervised loss that allows for learning hierarchical representations and have introduced RASA, a simple post-pretraining method to remove overtly spatial biases in the final representations. Across evaluations in image recognition, segmentation, robustness, OOD detection and 3D understanding, we find that it matches and frequently outperforms DINOv2 and other state-of-the-art models such as SigLIP 2. The recent release of DINOv3~\citep{simeoni2025dinov3} is a concurrent development. While it represents a significant advance, initial analysis (\autoref{fig:patch_entropy},~\autoref{fig:pca_viz}) suggests that positional bias remains a challenge, indicating the continued relevance of our technical contributions like RASA. Integrating our innovations into frameworks like DINOv3 is a promising direction for future work.

\section*{Acknowledgments.} 
This work was performed using HPC resources from GENCI-IDRIS (Grants \texttt{AD011015626}, \texttt{AD011016023}, \texttt{AD011016410}, \texttt{AD011016429}, \texttt{AD011016514}, \texttt{AD011016523}, \texttt{AD011016639}, and \texttt{AD011016537}). 
The authors gratefully acknowledge the scientific support and HPC resources provided by the Erlangen National High Performance Computing Center (NHR@FAU) of the Friedrich-Alexander-Universität Erlangen-Nürnberg (FAU) under the BayernKI project \texttt{v115be}. BayernKI funding is provided by Bavarian state authorities. Andrei Bursuc is partially funded by the
Horizon Europe project ELLIOT (GA No. 101214398).
We also thank Antonin Vobecky for his contribution to certain experiments, Rémi Lacroix for setting up the LAION-600M dataset and Monika Wysoczanska for her feedback on this paper.

{
    \small
    \bibliographystyle{ieeenat_fullname}
    \bibliography{main}
}

\clearpage

\newpage

\appendix

\newpage
\startcontents[app]

{\hypersetup{linkcolor=black}
  \section*{Contents}
  \printcontents[app]{l}{1}{}
}

\section{Related Work}
\label{sec:related_work}
Our work builds on and contributes to four major areas of prior research: self-supervised learning for visual representation learning, scaling strategies for data and model capacity in vision models, open-source foundation models, and techniques for disentangling semantic content from positional or representational biases.

\paragraph{Self-Supervised Learning (SSL) for Vision.}

Self-supervised learning
has emerged as a powerful paradigm for visual representation learning without any manual annotations. By designing pretext tasks that utilize image structure as supervision signals, SSL methods enable models to learn transferable features. Early approaches used handcrafted objectives such as context prediction~\citep{doersch2015unsupervised}, patch reordering~\citep{noroozi2016unsupervised}, colorization~\citep{zhang2016colorful,zhang2017split}, inpainting~\citep{pathak2016context}, geometric transformation prediction~\citep{gidaris2018unsupervised}, and instance discrimination~\citep{dosovitskiy2014discriminative, wu2018unsupervised}. 
Modern 
SSL methods primarily focus on learning invariances across augmented data views. While early approaches leverage contrastive learning \citep{oord2018representation,misra2020self,chen2020simple,he2020momentum,chen2020improved,chen2021empirical} by aligning positive pairs and separating negatives, bootstrap-based \citep{grill2020bootstrap,chen2021exploring, gidaris2021obow} and distillation-based methods \citep{caron2021emerging,oquab2024dinov2} refine targets through teacher-student networks, often removing the need for negative pairs. More recently, Masked Image Modeling (MIM) has emerged as a dominant SSL strategy, where models learn to reconstruct masked patches \citep{he2022masked,zhou2021ibot,kakogeorgiou2022hide,bao2022beit,wei2022masked}. Beyond these, clustering-based methods \citep{caron2018deep,caron2020unsupervised,ji2019invariant,sirko2025dip} have gained prominence, assigning pseudo-labels through algorithms like K-means or Sinkhorn-Knopp. The combination of MIM with clustering has shown particular promise, as exemplified by recent works such as MOCA \citep{gidaris2024moca} and CAPI \citep{darcet2025capi}. This hybrid approach leverages the strengths of both paradigms.

While current vision foundation models often fall into categories like vision-language or MAE-like architectures, which have their own strengths and limitations (e.g., reliance on text supervision or need for task-specific adaptations), DINOv2~\citep{oquab2024dinov2} stands out as a powerful pretrained model employing a clustering-based approach. However, DINOv2 has two key limitations: as a clustering method, it doesn't inherently capture the ambiguity often present in assignments at a fixed granularity, nor does it explicitly incorporate the benefits of modern hierarchical masking strategies. Our work 
addresses these concerns by integrating nested \emph{Matryoshka} projections \citep{kusupati2022matryoshka} directly into its objective. This allows each subspace to perform clustering at a \emph{different} granularity, yielding diverse pseudo-labels efficiently (see \autoref{fig:teaser}). Combined with an improved input masking strategy, our approach enables the joint learning of coarse-to-fine semantics without increasing model size, leading to strongly improved performances and reduction in memory.

\paragraph{Open Foundation Vision Models.}
The reliance on proprietary datasets in the training of current vision foundation models raises critical concerns regarding transparency, reproducibility, and the disentanglement of contributions. Models such as SEER~\citep{goyal2019scaling}, DINOv2~\citep{oquab2024dinov2}, CLIP~\citep{radford2021learning}, and billion-scale MAE~\citep{singh2023effectiveness} are all trained on proprietary data. This practice makes it challenging for the research community to isolate the true impact of model's novelty and training strategies from the unique characteristics and biases of the datasets themselves. The lack of access to these datasets hinders independent verification, fair comparison, and a comprehensive understanding of what truly drives model performance. 
Inspired by the success and large-scale utilization of CLIP, OpenCLIP~\citep{ilharco2021openclip, cherti2023reproducible} and MetaCLIP~\citep{xu2024demystifying, chuang2025meta} are the first to release models trained on public data. OpenCLIP~\citep{cherti2023reproducible, nezhurina2025scaling} stands out with the fully reproducible pipeline leveraging the ready-to-use LAION dataset~\citep{schuhmann2022laion, relaion}, study of scaling laws and release of intermediate checkpoints. 
On vision-only foundation models, the performance of DINOv2 has proven difficult to match by public data models.
Web-SSL~\citep{fan2025scaling} extends the study of large-scale self-supervised pretraining by training models on publicly available MetaCLIP-2B~\citep{xu2024demystifying} dataset, showing that models trained on open data can approach the performance of those trained on proprietary data on VLM tasks, but still below DINOv2 on visual perception tasks.

 Building on this, we present a fully open-source vision foundation model  using publicly available datasets, ReLAION~\citep{relaion}, as it represents the most popular and safe public dataset for large-scale vision model training.

\paragraph{Spatial correlations in learned representations.}
A common issue in dense self-supervised learning is the entanglement of semantic content with positional cues, causing models to rely on location rather than object identity. For instance, a model trained on “cows in grassy fields” and “camels in deserts” may misclassify a cow on a beach as a camel, due to learned associations with background context~\citep{arjovsky2019invariant}. Such spatial biases reduce generalization and can hinder performance when objects appear in atypical locations (e.g., a cow in the sky)~\citep{singh2020don}.

Several works have addressed this by proposing methods  invariant to positional information. \citet{lenc2015understanding} enforce equivariance to geometric transformations; \citet{wang2023projective} disentangle representations into orthogonal subspaces for content and style. Invariant Risk Minimization~\citep{arjovsky2019invariant} seeks features stable across environments, minimizing reliance on spurious cues. We propose a simple post-training strategy that learns a linear projection to identify and remove spatial information from features. Since, we use it as a post-training strategy, it requires no architectural changes and can be easily adapted to any pretrained model to reduce spatial bias.

\section{Implementation details}
\label{sec:app-setup}

A summary of the different datasets used at different training stage in given in~\autoref{tab:training-dataset}

\begin{table*}[h!]
\centering
\footnotesize
\begin{tabular}{lllr}
\toprule
\setlength{\tabcolsep}{4pt}
\textsc{Training type} & \textsc{Architecture} & \textsc{Dataset} & \textsc{\#Images} \\
\midrule
Pretraining & ViT-B & ImageNet-21K & 13,153,000 \\
Pretraining & ViT-L & LAION-COCO & 599,187,600 \\
Pretraining & ViT-G & LAION-COCO & 599,187,600 \\
\midrule
High-Res Finetuning & ViT-B, ViT-L & IN-1K, ADE20K, COCO, VOC, KITTI & 1,444,270 \\
\midrule
RASA Post-training & ViT-B/L/G & Pascal VOC & 17,125 \\
\bottomrule
\end{tabular}
\caption{Summary of training setup across architectures, datasets, and image counts.}
\label{tab:training-dataset}
\end{table*}

\paragraph{Pretraining Datasets}
We pretrain~\methodname on large-scale, publicly available image-only datasets to ensure full reproducibility. We use the ImageNet-21K~\citep{ridnik2021imagenet21k}, which contains approximately 13.1M high-quality, hierarchically labeled images across 21,841 classes. This dataset offers broad visual coverage and is widely used in foundation model pretraining. To further scale up training and improve generalization, we also leverage LAION-600M\footnote{\url{https://huggingface.co/datasets/laion/laion-coco}}, a subset of ReLAION-2B, which is a research-safe version of the LAION-5B dataset~\citep{schuhmann2022laion, relaion}. While LAION-5B is originally paired image-text data, we discard the text and use only the image modality.

\paragraph{Training}
~\methodname's architecture follows DINOv2~\citep{oquab2024dinov2} {without registers}, using Vision Transformers~\citep{dosovitskiy2020image} of varying model capacities: ViT-B with 86M parameters, ViT-L with 300M, and ViT-G with 1.1B. All models are trained from scratch for 625K iterations without distillation from larger models, unlike DINOv2, which distills from ViT-G into smaller variants. We use CyclicMask (pseudo-code in~\Cref{alg:cyclicmask}) and employ Matryoshka~\citep{kusupati2022matryoshka} with five nested heads with feature dimensions $[d, \frac{d}{2}, \dots, \frac{d}{16}]$ on top of the normal ViT backbone.

For LAION-600M, we use global crops scale of $[0.48, 1.0]$, following DINOv2-style augmentations. Stochastic depth regularization is set to 0.1 for ViT-B and 0.4 for ViT-L and ViT-G. We use a total batch size of 2048 for ViT-B and 3072 for both ViT-L and ViT-G, distributed across 32, 64 and 128 H100 GPUs for ViT-B, ViT-L and ViT-G respectively. The learning rate is set to $1 \times 10^{-3}$ for the Base model and $3.5 \times 10^{-4}$ for the Large and Giant variants, using a cosine schedule with warmup of 100K iterations.

We train RASA on top of our frozen backbone on Pascal VOC using crops of resolution $518 \times 518$, batch size of 128, with AdamW~\citep{loshchilov2017decoupled} optimizer. For each of the 8 incremental head training iterations, we used a dual-head linear projection (one for predictings and the other the y-axis patch positions) with sigmoid activation. In every iteration, only the top head was trained for 5 epochs with no weight decay and an initial learning rate of $2 \times 10^{-3}$, while all heads from previous iterations are absorbed into the weights of the final ViT layer, for iteratively removing the positional information from the features of the last ViT layer.

\begin{algorithm}[t]
\caption{Pseudo-code for \textsc{CyclicMask}}
\label{alg:cyclicmask}
\begin{algorithmic}[1]
\Require input size $(H, W)$, masking ratio $m \in [0,1]$, roll flag \texttt{roll}, 
        aspect ratio range $[r_{\min}, r_{\max}]$
\State $T \gets H \times W$ \Comment{total number of patches}
\State $n \gets \lfloor m \cdot T \rfloor$ \Comment{number of masking patches}

\State $c \gets \max(1,\, T - n)$ \Comment{number of complement patches}

\State $r \sim \exp\!(\mathcal{U}[\log r_{\min},\, \log r_{\max}])$ 
       \Comment{sample aspect ratio}

\State $h \gets \min\!(H,\, \lceil \sqrt{c \cdot r}\rceil)$
\State $w \gets \min\!(W,\, \lceil \sqrt{c / r}\rceil)$

\State \texttt{top} $\sim \mathcal{U}\{0,\, H - h\}$,\quad \texttt{left} $\sim \mathcal{U}\{0,\, W - w\}$

\State \texttt{mask} $\gets$ \texttt{torch.zeros(} $H$, $W$, \texttt{, dtype=bool)}
\State \texttt{mask[top:top + h, left:left + w]} $\gets$ \texttt{True}

\State \texttt{mask} $\gets$ $\lnot$\ \texttt{mask} \Comment{invert block to mask everything else}

\If{\texttt{roll} = \texttt{True}}
    \State \texttt{shift\_x} $\sim \mathcal{U}\{0,\, H-1\}$,\quad \texttt{shift\_y} $\sim \mathcal{U}\{0,\, W-1\}$
    \State \texttt{mask} $\gets$ \texttt{torch.roll(mask, (shift\_x, shift\_y), dims=(0,1))}
\EndIf

\State \Return \texttt{mask} of size $H \times W$
\end{algorithmic}
\end{algorithm}

\paragraph{High-resolution adaptation}
We initialize the model with pretrained weights and perform incremental high-resolution finetuning. First, we finetune the model at input resolution $364 \times 364$ with a local crop resolution of $112 \times 112$, using a base learning rate of $1.25 \times 10^{-5}$ for 30K iterations. The resulting checkpoint is then used to initialize the model for a second finetuning stage at input resolution $518 \times 518$ with a local crop size of $168 \times 168$, again with the same learning rate, for 10K iterations. All schedules are preserved but temporally compressed to fit within the shorter training horizons. The teacher network undergoes a warmup phase during the first 10K iterations of the initial $364$-resolution stage to stabilize early training dynamics. We use a dataset mix comprising only the training set of ImageNet-1K, ADE20K, COCO, KITTI, and VOC. Due to computational constraints, high-resolution finetuning is performed only for the ViT-B and ViT-L model.

\section{Evaluation details}
\paragraph{Overclustering}
In the overclustering setting, we follow the protocol of~\citep{ziegler2022self}. Specifically, we perform $K$-Means clustering (via faiss~\citep{johnson2019billion}) on the spatial tokens extracted from the backbone, discarding the projection head. To align clusters with ground-truth semantic labels, we first apply greedy matching based on pixel-level precision and then refine the assignment with Hungarian matching~\citep{kuhn1955hungarian}, ensuring permutation-invariant evaluation as in~\citep{ji2019invariant}. Inputs are cropped to 448$\times$448, while clustering operates on downsampled 100$\times$100 masks to reduce the computational cost of Hungarian matching. Results are reported as the mean Intersection-over-Union (mIoU), averaged over five seeds, across twp datasets: COCO-Thing, and Pascal VOC 2012~\citep{pascal-voc-2012}.

\paragraph{Visual In-Context Learning}  
The Dense Nearest Neighbor Retrieval Evaluation, introduced by~\citep{balazevic2024towards} and openly implemented by \citep{pariza2024hbird}, is designed to measure the scene understanding ability of dense image encoders through a retrieval-based protocol. The evaluation proceeds in three stages:  

\begin{enumerate}
    \item \textbf{Memory Bank Construction}: Given a training dataset with dense annotations, two memory banks are built. The first stores patch-level features obtained from the spatial outputs of a dense encoder, while the second stores the corresponding patch-level labels.  

    \item \textbf{Query Processing}: For each validation image, we extract patch embeddings from the encoder’s spatial output. Each query patch searches for its $k$ nearest neighbors in the feature memory bank, and the associated labels of these neighbors are aggregated to infer the query patch label.  

    \item \textbf{Evaluation}: After generating a predicted dense annotation for the full image, the result is compared against ground-truth labels to compute performance.  
\end{enumerate}  

Since the original implementation of~\citep{balazevic2024towards} is not publicly available, we rely on the open-source reimplementation from~\citep{pariza2024hbird}, which follows the authors’ description and allows leveraging either the ScaNN library~\citep{guo2020avq} or the faiss library ~\citep{johnson2019billion} for efficient nearest-neighbor search. In our experiments, we adhere closely to this setup but make two modifications: (1) instead of restricting memory to a fixed capacity of 10,240,000 entries, we index all patch embeddings extracted from images resized to $518 \times 518$; and (2) we employ GPU-accelerated FAISS for nearest-neighbor retrieval, configured to approximate the ScaNN setup used in the original evaluation (e.g., with $k=30$ neighbors).  

We report results as mean Intersection-over-Union (mIoU) on two benchmark datasets: Pascal VOC 2012~\citep{pascal-voc-2012} and ADE20K~\citep{zhou2017scene}, averaging over five random seeds. 

\paragraph{Linear Segmentation}
Linear segmentation is a common protocol to assess the linear separability of learned spatial representations. 
The setup freezes the encoder and trains a lightweight segmentation head, typically a single linear layer with batch normalization, using pixel-wise cross-entropy loss. 
In practice, the spatial patch embeddings produced by the backbone are bilinearly upsampled to match the resolution of the ground-truth masks, and the linear head is trained on top. 
This procedure is implemented in \texttt{mmsegmentation} and was used in the DINOv2 paper~\citep{oquabdinov2} for evaluating Pascal VOC and ADE20K, where the backbone is kept frozen and only the linear head is optimized.

In this paper, we adopt the DINOv2 setup as a baseline and introduce a few modifications tailored to our experimental needs:

\begin{itemize}
    \item \textbf{General changes (both datasets):}  
    We switch from iteration-based training (\texttt{max\_iters = 40000}) to epoch-based training with a maximum of 50 epochs. 
    The crop size is increased from $512 \times 512$ to $518 \times 518$, and the stride is adjusted accordingly to half the crop size. 
    Data augmentation uses \texttt{RandomResize} instead of \texttt{Resize}. 
    The learning rate schedule is also defined by epoch rather than iteration.

    \item \textbf{Pascal VOC 2012~\citep{pascal-voc-2012}:}  
    We replace the AdamW optimizer (used in DINOv2) with SGD (lr = 0.001, momentum = 0.9, weight decay = $5\times10^{-4}$). 
    The resize target is set to $(2048, 518)$ with a ratio range of $(0.5, 2.0)$, consistent with the crop size change.

    \item \textbf{ADE20K~\citep{zhou2017scene}:}  
    We keep AdamW as the optimizer but wrap it with an \texttt{OptimWrapper}. 
    The random resize ratio range is expanded from $(0.5, 2.0)$ to $(1.0, 3.0)$, making the scale augmentation more aggressive. 
    At test time, instead of single-scale evaluation, we adopt multi-scale testing with image ratios $[1.0, 1.32, 1.73]$. 
\end{itemize}

These modifications preserve the core linear segmentation evaluation protocol of DINOv2 while adapting it for our framework and experimental goals. We generally have seen the modifications to work better and improve further the results of all the models we tested.

\paragraph{Evaluation Datasets}
\label{appendix:eval_datasets}

\textbf{Pascal VOC 2012} \citep{pascal-voc-2012}.
We use the most recent \texttt{trainaug} split, which contains 10,582 annotated images across 21 categories, including one background class. The validation set includes 1,449 images. We exclude unlabeled objects and the boundary class from evaluation. For hyperparameter tuning of the fully unsupervised segmentation method of~\citep{ziegler2022self}, we rely on the original \texttt{train} split (1,464 images).

\textbf{COCO-Stuff 164K} \citep{caesar2018cocostuff}.
The full COCO-Stuff dataset provides semantic annotations for both “stuff” (background, amorphous regions) and “thing” (foreground, countable objects) categories, with 91 stuff and 80 thing classes, respectively. It contains 118,000 training images and 5,000 validation images.
In prior work~\citep{ziegler2022self,pariza2025near}, two variants of this dataset are considered:
\begin{enumerate}
    \item \emph{COCO-Stuff}, where the 91 stuff categories are grouped into 15 broader classes (with all thing categories collapsed into a single “other” label that is ignored during evaluation).
    \item \emph{COCO-Thing}, where the 80 thing categories are consolidated into 12 broader object classes, and background regions are excluded from evaluation.
\end{enumerate}

In this work, we only make use of the \emph{COCO-Thing} variant. Specifically, we follow~\citep{kirillov2019panoptic} to obtain panoptic instance annotations, which are then merged into object-level categories using the official conversion script. The resulting 12-class setup emphasizes object-level reasoning in cluttered natural scenes and serves as our benchmark for overclustering experiments. Although we refer to the dataset as “COCO-Stuff 164K” for consistency with the literature, only the COCO-Thing portion is used in our evaluations.

\textbf{ADE20K} \citep{zhou2017scene}.
ADE20K is a large-scale scene parsing benchmark with 150 semantic categories, ranging from background classes (e.g., sky, grass) to fine-grained objects (e.g., person, car). The dataset contains 20,210 training images and 2,000 validation images with detailed annotations. Due to its diversity and fine-grained structure, ADE20K is widely regarded as one of the most challenging datasets for dense prediction tasks. In our experiments, we use the full dataset but ignore the \textit{others} label during evaluation.

\section{Additional Results}

\begin{figure}[t]
    \centering
    \includegraphics[width=1.0\linewidth]{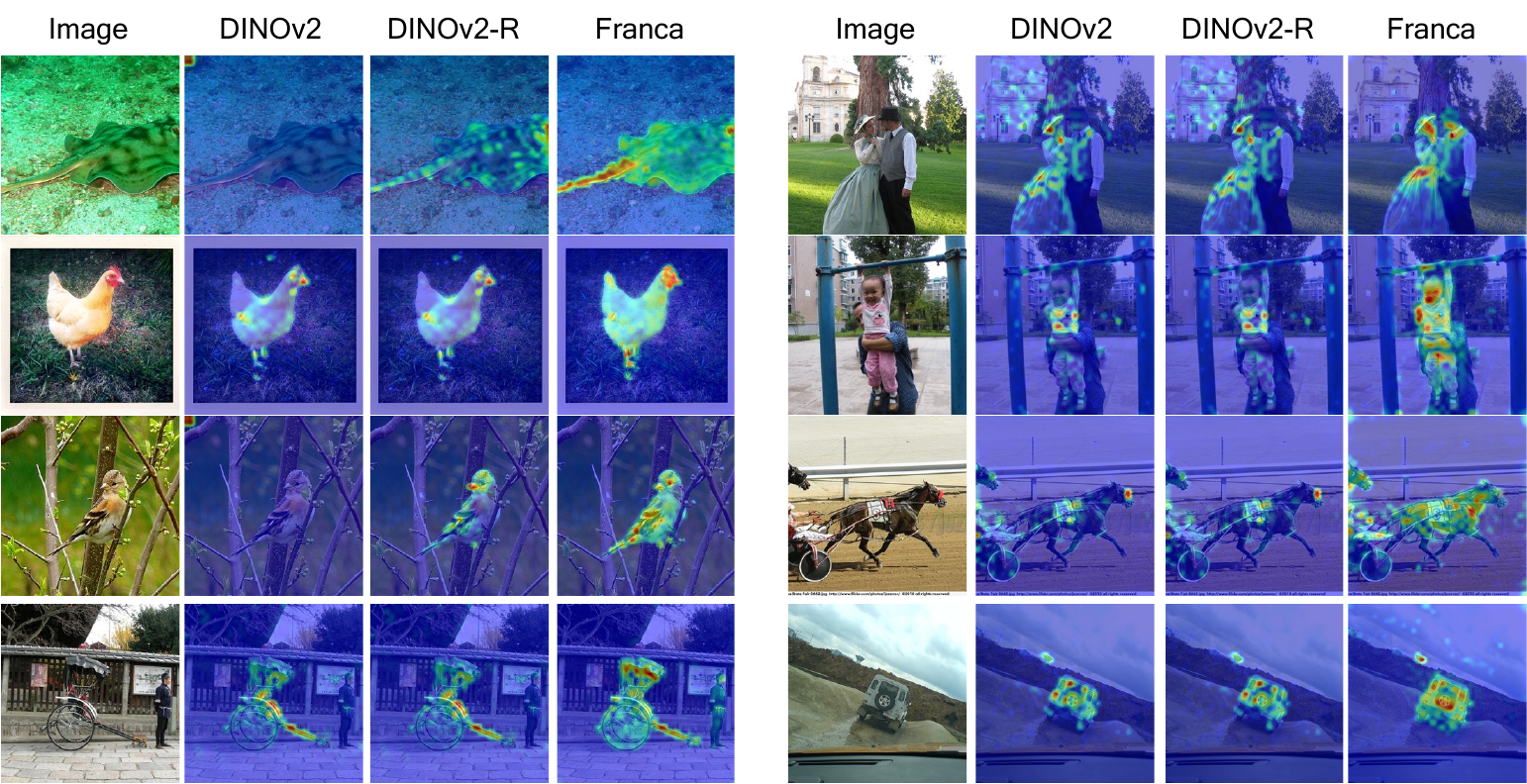}
    \caption{\emph{Self-attention maps} utilizing  $14 \times 14$ patches. These maps are visualized using the \texttt{[CLS]} token on the last layer's heads on the validation set of ImageNet-1K~\citep{russakovsky2015imagenet}. ~\methodname has better localization than DINOv2 with registers~\citep{darcet2023vision} without requiring the use of registers, where the nested Matryoshka clustering captures fine-grained details, e.g., feathers, beaks of bird.}
    \label{fig:attention-viz}
\end{figure}

\alert{
\subsection{Training with registers}

Registers mitigate feature artifacts (high-norm outlier tokens) in DINOv2. However, Franca's Nested Matryoshka Clustering intrinsically prevents these artifacts by enforcing multi-scale semantic structure, evidenced by lower token norms in \autoref{fig:avg-token-norm} compared to even DINOv2-R. Consequently, adding registers to Franca is redundant and subpar, degrading k-NN on IN-1K (79.6$\rightarrow$78.7) and linear segmentation on ADE20K (46.2$\rightarrow$45.5). As registers primarily benefit DINOv2's dense tasks without improving classification, and Franca naturally yields coherent dense features as shown in ~\autoref{fig:pca_viz} and sharp attention maps (\autoref{fig:attention-viz}), we use DINOv2 \emph{without} registers (except Tab 2(b)) as the primary baseline to ensure a fair comparison. 

\begin{figure}[h]
\centering
    \includegraphics[width=\linewidth]{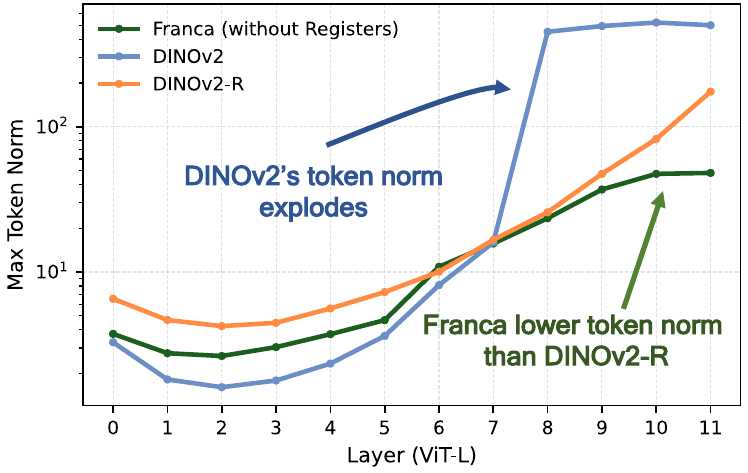}
    \caption{Layer-wise maximum token norms of Franca (LAION-600M), DINOv2 (LVD-142M) and DINOv2-R (LVD-142M) with ViT-L backbone on validation set of ImageNet-1K.}
\label{fig:avg-token-norm}
\vspace{-13pt}
\end{figure}
}

\begin{table*}[ht]\centering
\footnotesize
\begin{tabular}{llcccccccccccc}
\toprule
Method               & Arch       & Food   & C10   & C100   & SUN   & Cars   & Aircr  & DTD   & Pets  & Cal    & Flwrs & CUB  & Avg   \\
\midrule
MAE                  & ViT-H/14   & 78.4   & 96.1  & 83.9   & 63.9  & 56.1   & 63.4   & 75.4   & 89.4  & 95.9   & 92.3    & 57.2   & 77.5   \\
DINOv2$^\dagger$     & ViT-L/14   & 93.4   & 99.2  & 93.9   & 78.1   & 89.9   & 81.7   & 82.9  & 95.2 & 87.2  & 99.6    & 90.3    & 90.1     \\
DINOv2$^\S$          & ViT-L/14   & 94.3  & 99.3  & 93.4   & 78.7  & 89.9   & 81.5   & 84.0  & 96.5   & 97.5    & 99.7    & 90.5  & 91.4   \\
Web-SSL              & ViT-L/14   & 91.0   & 98.9  & 90.7   & 77.5  & 88.9   & 80.2   & 83.6  & 93.1  & 95.1   & 98.8    & 90.9   & 89.9   \\
\cmidrule{2-14} 
DINOv2               & ViT-G/14   & 94.7   & \textbf{99.5}  & 94.4  & 78.7   & 91.4   & \textbf{87.2}   & 84.5  & 96.7  & \textbf{97.6}   & \textbf{99.7}    & \textbf{91.6}   & \textbf{92.3}   \\
Web-SSL              & ViT-G/14   & 94.1   & 99.4  & 93.1   & 78.0  & 90.3   & 83.7   & 84.7  & 92.4  & 96.8   & 99.4    & 91.2  & 91.2   \\
OpenCLIP             & ViT-G/14   & 94.5   & 98.7  & 91.0   & 84.0  & \textbf{96.1}   & 80.2   & \textbf{86.0}  & 95.7  & 98.1   & 99.5    & 89.9  & 92.2   \\
\midrule 
\rowcolor{rowmint}
\methodname (ours)   & ViT-B/14   & 90.6   & 98.7  & 90.9   & 77.0  & 88.7   & 75.2   & 81.7  & 94.1  & 96.2   & 99.7    & 86.2  & 88.9   \\
\rowcolor{rowmint}
\methodname (ours)   & ViT-L/14   & 94.3   & 99.4  & 94.1   & 79.9  & 89.5   & 81.3   & 84.1  & 95.1  & 97.4   & 99.8  & 91.1   & 91.5        \\
\rowcolor{rowmint}
\methodname(ours) & ViT-G/14   & \textbf{95.0}   & \textbf{99.5}  & \textbf{95.1}   & \textbf{78.9}  & 91.3   & 85.5   & 85.0  & \textbf{97.2}  & 97.5   & \textbf{99.7}    & 91.3  & \textbf{92.3}   \\

\bottomrule
\end{tabular}
\vspace{3pt}
\caption{\emph{Linear evaluation of frozen features on fine-grained datasets}. top-1 accuracy measured across 11 benchmarks across objects, scenes, and textures, following ~\citep{chen2020simple}; $^\dagger$: reproduced on IN-21K without distillation; $^\S$: distilled from DINOv2-G on LVD-142M. 
}
\label{tab:vision-benchmark}
\end{table*}

\subsection{Fine-grained classification}
We evaluate the transferability of the learned representations on 11 classification benchmarks introduced in SimCLR~\citep{chen2020simple}. These benchmarks cover a variety of tasks, including scene recognition, fine-grained object classification (such as food, cars, and aircraft), and texture recognition. Following~\citep{oquab2024dinov2}, we train a logistic regression classifier on features extracted from a frozen backbone. This approach focuses solely on the quality of the visual features and provides a fair way to compare performance across different tasks. Although some of these benchmarks tend to favor models trained with text supervision, our features perform strongly and competitively across many categories.

As shown in~\autoref{tab:vision-benchmark}, our method, \methodname, transfers well across a wide range of downstream tasks. Our ViT-G/14 model (\methodname-G) achieves the same performance as DINOv2-G and outperforms Web-SSL-G by 1.1\% despite being trained on much less data. It also matches the performance of larger models like OpenCLIP-G. On datasets such as CIFAR-100 and Oxford Pets,~\methodname-G achieves 0.7\% and 0.5\% gains over DINOv2G respectively, demonstrating \methodname's strong generalization ability across both natural and fine-grained classification tasks.

\alert{
\subsection{Computational Cost}
We report the computational cost in~\autoref{tab:compute-cost} providing estimates that cover the full training pipeline, including finetuning at $364\times364$ and $518\times518$ resolutions. Compute cost scales roughly linearly with model size: \methodname-G requires about 13$\times$ the GPU-hours of \methodname-B, while \methodname-L lies between at 4$\times$. This reflects both the higher per-step cost of larger models and the increased GPU parallelism needed to maintain throughput. Compared to prior large-scale self-supervised training efforts such as MetaCLIP and DINOv2-G, our models use 
fewer resources, \methodname-G trains in $\sim$20K GPU-hours on H100s, over an order of magnitude less than MetaCLIP’s 368K GPU-hours for a similar training horizon.

Reported numbers include practical overheads (e.g., hyperparameter tuning, checkpoint restarts, failed runs), providing a realistic measure of total compute. High-resolution finetuning accounts for a notable portion of the cost, especially for \methodname-B and \methodname-L, underscoring the need to balance resolution scaling with available compute.

\begin{table}[h]
\centering
\footnotesize
\setlength{\tabcolsep}{2pt}
\begin{tabular}{lcccc}
\toprule
Model Arch. & \# GPUs & Trainable & Iterations & GPU hours \\
            &         & Params (in M)    & (Pretrain$+$ HRFT)  & \\
\midrule
DINOv2-B    & 32      & 184       & 625K + 20K  & 1,324  \\
Franca-B    & 32     & 236        & 625K + 40K  & 1,504  \\
\cmidrule{2-5}
DINOv2-L    & 64      & 404       & 625K + 20K  & 5,677  \\
Franca-L    & 64     & 457        & 625K + 40K  & 5,952 \\
\cmidrule{2-5}
MetaCLIP-G  & 256     & 1800      & 390K        & 368,640 \\
DINOv2-G    & 128     & 1240      & 625K + 20K  &  22,016 \\
Franca-G    & 128    & 1295       & 625K        & 19,992 \\
\bottomrule
\end{tabular}
\vspace{-3pt}
\caption{We estimate the overall GPU hours for our project, including hyper-parameter search, restarts from checkpoints, and failed runs, amounting to a total of 160K GPU-hours on NVIDIA H100 GPUs (80GB, NVLink interconnect). The table provides a breakdown across model scales (ViT-B, ViT-L, ViT-G) and training phases (pretraining vs. finetuning). Reported values reflect actual wall-clock consumption rather than idealized FLOP counts.}
\label{tab:compute-cost}
\vspace{-6pt}
\end{table}

}

\alert{
\subsection{Effectiveness of slicing features}

We compare both Franca-L (LAION-600M) and DINOv2-L (LVD-142M) features after applying PCA and slicing the first $k$ components. Linear probing on IN-1K in ~\autoref{tab:rebut-slice} shows DINOv2-L degrades quickly (86.3 $\rightarrow$ 22.0 from 1024 $\rightarrow$ first 50 components), while Franca remains more robust (retaining 84.9 $\rightarrow$ 37.2), showing that Matryoshka preserves useful structure without post-hoc dimensionality reduction.

\begin{table}[h]
\centering
\small
\setlength{\tabcolsep}{3pt}
\begin{tabular}{lccccc}
\toprule
\textsc{:dim (after PCA)} & $200$ & $100$ & $50$ & $25$ & $10$ \\
\midrule
DINOv2-L        & 71.2       & 48.6      & 22.8    & 8.7     & 2.0         \\
\rowcolor{rowmint}
Franca-L        & 70.1       & 54.4      & 37.2    & 23.4    & 12.8         \\
\bottomrule
\end{tabular}
\caption{Franca encodes better semantics across the sliced features as compared to DINOv2.}
\label{tab:rebut-slice}
\end{table}
}

\subsection{Attention maps visualizations}
We compare self-attention maps from the final layer's \texttt{[CLS]} token of DINOv2, DINOv2R (with registers)~\citep{darcet2023vision} and~\methodname in~\autoref{fig:attention-viz}. DINOv2 often fails to localize objects, especially under clutter or occlusion, while DINOv2R offers only minor improvements. In contrast,~\methodname yields sharply focused attention maps aligned with object boundaries, even for small or partially occluded instances. This suggests that our Matryoshka-style multi-head clustering promotes semantically rich features and finer-grained representations.

\begin{table*}[t]
\centering
\footnotesize
\setlength{\tabcolsep}{5pt}
\begin{tabular}{lllcccc|c}
\toprule
\textsc{Model} & \textsc{Arch.} & \textsc{Data} & $d{=}0$ & $d{=}1$ & $d{=}2$ & All & \textsc{SI-RMSE $\downarrow$} \\
\midrule
\multicolumn{8}{c}{\textsc{Image resolution $800 \times 800$ (SPair-71K) and $480 \times 480$ (NYUv2)}} \\
\midrule
OpenCLIP         & ViT-B/14  & LAION-2B    & 18.31 & 16.78 & 17.05 & 17.63 & 0.39 \\
SigLIP2          & ViT-B/14  & WebLI       & 9.29  & 5.96  & 5.82  & 7.83  & 0.56 \\ 
DINOv2$^\dagger$ & ViT-B/14  & IN-21K      & 42.82 & 33.38 & 34.82 & 38.71 & 0.33 \\
DINOv2$^\S$      & ViT-B/14  & LVD-142M    & 63.04          & \textbf{51.09} & \textbf{48.76} & \textbf{55.98} & \textbf{0.25} \\
\rowcolor{rowmint}
\methodname (ours) & ViT-B/14 & IN-21K     & 53.46 & 41.9 & 44.2 & 45.6 & 0.30 \\
\addlinespace
\cmidrule{2-8}
OpenCLIP         & ViT-L/14 & LAION-2B      & 22.73  & 20.18  & 20.37 & 21.26  & 0.37  \\
SigLIP2          & ViT-L/14 & WebLI         & 8.66  & 5.38    & 5.05  & 6.88   & 0.55 \\
DINOv2$^\dagger$ & ViT-L/14 & IN-21K        & 57.60 & 42.91 & 44.17 & 50.68 & 0.31 \\
DINOv2$^\S$      & ViT-L/14 & LVD-142M      & 63.91         & \textbf{53.11} & \textbf{51.91}  & 56.92           & \textbf{0.23} \\
\rowcolor{rowmint}
\methodname (ours) & ViT-L/14 & LAION-600M  & 58.50          & 46.74           & 48.28            & 51.76            & 0.25 \\
\bottomrule
\end{tabular}

\caption{\emph{Probing 3D understanding} via keypoint matching on SPair-71K and monocular depth estimation on NYUv2. We report correspondence accuracy (higher is better) under increasing viewpoint changes ($d{=}0, 1, 2$, and overall) on SPair-71K, and scale-invariant RMSE (lower is better) for depth estimation on NYUv2. Results are shown for high ($800^2$) resolution SPair-71K input. $^\dagger$: reproduced on IN-21K without distillation; $^\S$: distilled from DINOv2-G. %
}
\label{tab:3d_understanding}
\end{table*}

\subsection{Probing 3D understanding}

We evaluate the geometric understanding of~\methodname on two tasks: keypoint correspondence (SPair-71k~\citep{min2019spair}) and monocular depth estimation (NYUv2~\citep{silberman2012indoor}). For SPair-71k, the goal is to establish dense keypoint correspondences under varying viewpoint changes, where we report accuracy across all keypoints and under increasing viewpoint disparity. For depth estimation, we follow the AdaBins protocol~\citep{bhat2021adabins} and measure performance using the scale-invariant root-mean-square error (SI-RMSE). All evaluations are conducted at high resolution: $800 \times 800$ for SPair-71k and $480 \times 480$ for NYUv2.

As shown in~\autoref{tab:3d_understanding},~\methodname achieves strong performance across both tasks. On SPair-71k, it outperforms DINOv2 trained on the ImageNet-21K dataset, both at ViT-B and ViT-L models. 
On NYUv2,~\methodname achieves comparable performance as DINOv2 distilled from larger DINOv2 model. This is also due to the fact that DINOv2 uses NYU Depth v2 during pretraining (as part of LVD 142M).  These results indicate that~\methodname learns a spatial representation that captures both fine-grained 2D alignment and coarse 3D structure, generalizing well from object-centric pretraining to downstream geometric tasks.

\subsection{Overclustering}
\begin{table}[h]
\centering

\footnotesize
\setlength{\tabcolsep}{3pt}
\begin{tabular}{ll|cccc}
\toprule
\multirow{2}{*}{\textsc{Method}} & \multirow{2}{*}{\textsc{Backbone}} & \multicolumn{4}{c}{\textsc{OverClustering}} \\
\cmidrule{3-6}
                  &            & \multicolumn{2}{c}{\textsc{VOC}} & \multicolumn{2}{c}{\textsc{COCO-Things}}\\
\cmidrule{3-4} \cmidrule{5-6}
                  &            & $K=100$ & $K=300$ & $K=100$ & $K=300$ \\
\midrule
SigLIP            & ViT-B/16   & 29.5    & 36.9      & 41.5      & 53.4       \\
iBOT              & ViT-B/16   & 21.2    & 29.1      & 18.3      & 26.3        \\
EVA-CLIP          & ViT-B/16   & \textbf{43.3}    & 49.1      & 41.3      & 52.0       \\
DINOv2$^\dagger$  & ViT-B/14   & 25.9    & 34.7      & 20.5      & 28.7         \\
DINOv2$^\S$       & ViT-B/14   & 39.2    & 52.5      & \textbf{46.5}      & \textbf{54.0 }        \\
\cmidrule{2-6}
\rowcolor{rowmint}
~\methodname (ours) & ViT-B/14   & 37.5    & \textbf{56.4}      & 38.8      & 51.1        \\
\midrule
SigLIP 2          & ViT-L/16   & 24.3    & 40.4      & 43.5      & 50.7       \\
Web-SSL           & ViT-L/14   & 28.2    & 37.7      & 26.3      & 33.1        \\
DINOv2$^\dagger$  & ViT-L/14   & 25.9    & 34.7      & 24.1      & 35.1         \\
DINOv2$^\S$       & ViT-L/14   & 26.5    & 43.0      & 34.8      & 45.7       \\
\cmidrule{2-6}
\rowcolor{rowmint}
~\methodname (ours) & ViT-L/14   & \textbf{47.4}    & \textbf{58.9}      & \textbf{49.6}      & \textbf{54.4}        \\
\midrule
Web-SSL           & ViT-G/14   & 26.0    & 33.4           & 15.5 &     21.5      \\
DINOv2            & ViT-G/14   & 19.5    & 27.7         & 20.7    & 29.2    \\
\cmidrule{2-6}
\rowcolor{rowmint}
~\methodname (ours) & ViT-G/14   & \textbf{39.4 }   & \textbf{49.2}            & \textbf{25.8} & \textbf{31.9}         \\
\bottomrule
\end{tabular}
\vspace{6pt}
\caption{\emph{OverClustering Performance.} Comparison of overclustering performance (mIoU) on Pascal VOC and COCO-Things datasets with $K=100$ and $K=300$. $^\dagger$: reproduced on IN-21K, without distillation; $\S$ : distilled from DINOv2-G on LVD-142M.}
\label{tab:overclustering_results}
\end{table}

We evaluate \methodname using the overclustering protocol from~\citep{ziegler2022self}, which measures semantic alignment of spatial features in a label-free setting. Patch embeddings are clustered with $K$-Means and matched to ground-truth segmentation masks via Hungarian matching~\citep{kuhn1955hungarian}, and performance is reported as mean IoU (mIoU). This task highlights the ability of representations to capture fine-grained structure, which is crucial for dense prediction tasks such as semantic segmentation and object detection.

As shown in~\autoref{tab:overclustering_results}, \methodname consistently outperforms strong baselines across backbones and clustering granularities. On ViT-B/14, it achieves the highest VOC performance at $K=300$ (56.4 mIoU), surpassing DINOv2-B$^\S$ (52.5) while remaining competitive on COCO-Things. With ViT-L/14, \methodname delivers its strongest results: it reaches 47.4 and 58.9 mIoU on VOC ($K=100$/$K=300$), and 49.6 and 54.4 on COCO-Things, outperforming DINOv2-L$^\S$, Web-SSL, and SigLIP 2. At ViT-G/14 scale, \methodname maintains a clear lead over DINOv2-G and Web-SSL, reaching 49.2 on VOC ($K=300$) and 31.9 on COCO-Things ($K=300$), despite the challenging scaling regime.

These results underscore that \methodname not only scales effectively to larger backbones but also delivers robust gains on both coarse (VOC) and complex (COCO-Things) datasets. Importantly, it rivals or surpasses multimodal baselines like EVA-CLIP and SigLIP 2, highlighting its strength as a unimodal self-supervised learner for dense prediction settings.

\subsection{Ablating RASA }
\label{app:rasa_ablations}
\subsubsection{Dataset Size Requirements for Training the RASA Head}
\label{app:rasa_dataset_size}
\begin{table*}[t]
\caption{\emph{Ablating the Dataset Size used for Training RASA Head}}
\setlength{\tabcolsep}{3pt}
\centering
\begin{subtable}[t]{0.48\textwidth}
\centering
\begin{tabular}{lrrrrr}
\toprule
\multicolumn{1}{c}{} & \multicolumn{2}{c}{\textsc{In-context}} & \multicolumn{2}{c}{\textsc{Lin. Seg.}} \\
\cmidrule(lr){1-1}\cmidrule(lr){2-3}\cmidrule(lr){4-5}
Fraction & VOC  & ADE20K  & VOC  & ADE20K \\
\midrule
0.1 & 76.6 & 35.2 & 89.3 & 46.2 \\
0.2 & \textbf{76.7} & \textbf{35.3} & \textbf{89.4} & \textbf{46.2} \\
0.3 & 76.7 & 35.3 & 89.3 & 46.0 \\
0.4 & 76.7 & 35.3 & 89.3 & 46.1 \\
0.5 & 76.7 & 35.3 & 89.3 & 46.2 \\
0.6 & 76.7 & 35.3 & 89.3 & 46.1 \\
0.8 & 76.7 & 35.3 & 89.3 & 46.2 \\
\bottomrule
\end{tabular}
\caption{Fractions of the COCO Dataset}
\label{tab:ablations_rasa_data_left}
\end{subtable}
\begin{subtable}[t]{0.48\textwidth}
\centering
\begin{tabular}{lrrrrr}
\toprule
\multicolumn{1}{c}{} & \multicolumn{2}{c}{\textsc{In-context}} & \multicolumn{2}{c}{\textsc{Lin. Seg.}} \\
\cmidrule(lr){1-1}\cmidrule(lr){2-3}\cmidrule(lr){4-5}
Fraction & VOC  & ADE20K  & VOC  & ADE20K \\
\midrule
0.1 & \textbf{76.7} & 35.3 & \textbf{89.4} & \textbf{46.1} \\
0.2 & 76.7 & \textbf{35.4} & 89.4 & 45.9 \\
0.3 & 76.7 & 35.4 & 89.3 & 46.0 \\
0.4 & 76.7 & 35.4 & 89.4 & 46.0 \\
0.5 & 76.7 & 35.4 & 89.4 & 45.9 \\
0.6 & 76.7 & 35.4 & 89.3 & 46.0 \\
0.8 & 76.7 & 35.4 & 89.4 & 45.9 \\
\bottomrule
\end{tabular}
\caption{Fractions of the Imagenet Dataset}
\label{tab:ablations_rasa_data_right}
\end{subtable}
\label{tab:ablations_rasa_data}
\end{table*}

We ablate the dataset size required for training the \textbf{RASA} head by varying the fraction of images sampled from \textsc{COCO} (scene-centric) and \textsc{ImageNet100} (object-centric). Tables~\ref{tab:ablations_rasa_data_left} and \ref{tab:ablations_rasa_data_right} summarize the results across in-context segmentation and linear segmentation.

\vspace{0.5em}
\noindent\textbf{Observations.}  
On \textsc{COCO}, as little as $10\%$--$20\%$ of the data suffices to reach peak performance, with VOC and ADE20K scores saturating at $76.6$--$76.7$ and $35.2$--$35.3$ (in-context), and $89.3$--$89.4$ and $46.2$ (linear segmentation). Increasing the fraction up to $80\%$ yields no further improvements, indicating that performance plateaus once $\sim$10k images are used.

A similar trend is observed on \textsc{ImageNet100}, where $10\%$--$20\%$ of the dataset again provides optimal results ($76.7$ VOC / $35.3$--$35.4$ ADE20K in-context, $89.4$ VOC / $45.9$--$46.1$ ADE20K for linear segmentation). Larger fractions show negligible variation, well within noise levels.

\vspace{0.5em}
\noindent\textbf{Conclusion.}  
These findings demonstrate two key points. First, training the RASA head is highly \emph{data-efficient}: roughly $10$k images are sufficient to achieve strong disentanglement. Second, the type of dataset---\textsc{COCO} (scene-oriented) vs. \textsc{ImageNet100} (object-oriented)---does not materially affect downstream performance. This indicates that RASA is largely agnostic to the semantic bias of the training distribution and can be reliably trained with minimal data.  

Based on these observations, we adopt \textsc{Pascal VOC} as our default choice for training the RASA head, since it is compact yet sufficient to reach optimal performance. With this setup, we achieve strong results across multiple benchmarks (Table~\ref{tab:rasa_voc_results}). This confirms that a lightweight dataset such as Pascal VOC is adequate for effective training, while maintaining competitive performance across diverse tasks. 

\begin{table}[h]
\centering
\caption{Training the RASA head on top of \textsc{ViT-Base} of ~\methodname on Pascal VOC.}
\label{tab:rasa_voc_results}
\setlength{\tabcolsep}{6pt}
\begin{tabular}{lcccc}
\toprule
\multirow{2}{*}{\textsc{Setup}} & \multicolumn{2}{c}{\textsc{In-context}} & \multicolumn{2}{c}{\textsc{Lin. Seg.}} \\
\cmidrule(lr){2-3}\cmidrule(lr){4-5}
& VOC & ADE20K & VOC & ADE20K \\
\midrule
~\methodname & 76.2 & 35.0 &  89.4 & 46.0 \\
~\methodname + \textsc{RASA} & \textbf{76.7} & \textbf{35.3} & \textbf{89.4} & \textbf{46.0} \\
\bottomrule
\end{tabular}
\end{table}

\subsubsection{Learning Rate for Training the Dual-Linear Position Predictor}
\label{app:rasa_lr}
\begin{table*}[t]
\caption{\emph{Ablating the learning rate used to train each dual linear position predictor.}}
\setlength{\tabcolsep}{3pt}
\centering
\begin{subtable}[t]{0.48\textwidth}
\centering
\begin{tabular}{lrrrrr}
\toprule
\multicolumn{1}{c}{} & \multicolumn{2}{c}{\textsc{In-context}} & \multicolumn{2}{c}{\textsc{Lin. Seg.}} \\
\cmidrule(lr){1-1}\cmidrule(lr){2-3}\cmidrule(lr){4-5}
lr & VOC  & ADE20K  & VOC  & ADE20K \\
\midrule
0.002 & \textbf{76.7} & \textbf{35.3} & \textbf{89.4 }& \textbf{46.0} \\
0.005 & 76.7 & 35.3 & 89.4 & 45.7 \\
0.0001 & 76.3 & 34.7 & 89.4 &  45.8 \\
0.0005 & 76.6 & 35.2 & 93.9 &  46.0 \\
\bottomrule
\end{tabular}
\caption{Learning Rates on the Pascal VOC}
\label{tab:ablations_rasa_lr_left}
\end{subtable}
\begin{subtable}[t]{0.48\textwidth}
\centering
\begin{tabular}{lrrrrr}
\toprule
\multicolumn{1}{c}{} & \multicolumn{2}{c}{\textsc{In-context}} & \multicolumn{2}{c}{\textsc{Lin. Seg.}} \\
\cmidrule(lr){1-1}\cmidrule(lr){2-3}\cmidrule(lr){4-5}
lr & VOC  & ADE20K  & VOC  & ADE20K \\
\midrule
0.002 & 76.7 & 35.3 & 89.4 &  45.9 \\
0.005 & 76.5 & 35.0 & 89.4 &  46.1 \\
0.0001 & \textbf{76.7} & \textbf{35.3 }& \textbf{89.4 }& \textbf{46.2} \\
0.0005 & 76.6 & 35.2 & 89.4  & 45.9 \\
\bottomrule
\end{tabular}
\caption{Learning Rates on the COCO Dataset}
\label{tab:ablations_rasa_lr_right}
\end{subtable}
\label{tab:ablations_rasa_lr}
\end{table*}

We also ablate the learning rate used when training each dual-linear position predictor in the RASA head. Results are reported on both \textsc{Pascal VOC} and \textsc{COCO} in Tables~\ref{tab:ablations_rasa_lr_left} and \ref{tab:ablations_rasa_lr_right} respectively.  

\vspace{0.5em}
\noindent\textbf{Observations.}  
On \textsc{Pascal VOC} in \autoref{tab:ablations_rasa_lr_left}, the optimal performance is achieved with a learning rate of $0.002$, reaching $76.7$ VOC / $35.3$ ADE20K for in-context and $89.4$ VOC / $46.0$ ADE20K for linear segmentation. Larger learning rates such as $0.005$ yield slightly reduced segmentation accuracy, while very small rates ($0.0001$) underperform across both tasks. Interestingly, $0.0005$ produces a competitive ADE20K score ($46.0$) but is less stable overall.  

On \textsc{COCO} in \autoref{tab:ablations_rasa_lr_right}, the trend differs: the best results are obtained at the much smaller learning rate of $0.0001$ ($76.7$ VOC / $35.3$ ADE20K in-context, $89.4$ VOC / $46.2$ ADE20K linear segmentation). Higher learning rates ($0.002$, $0.005$) still maintain strong performance but introduce minor drops or fluctuations, particularly in VOC segmentation. The consistency of $0.0001$ across all metrics suggests that COCO benefits from more conservative updates during training.  

\vspace{0.5em}
\noindent\textbf{Conclusion.}  
These results indicate that while the RASA head is generally robust to a wide range of learning rates, the optimal configuration is dataset-dependent. For \textsc{Pascal VOC}, a moderately large learning rate ($0.002$) works best, while for \textsc{COCO}, a smaller learning rate ($0.0001$) provides the most stable and accurate results. Overall, this highlights that tuning the learning rate can provide small but measurable gains, though the model remains relatively stable across settings.

\subsubsection{Number of Epochs for Training the Dual-Linear Position Predictor}
\label{app:rasa_epochs}
\begin{table*}[t]
\caption{\emph{Ablating the number of Epochs used to train each dual linear position predictor.}}
\setlength{\tabcolsep}{3pt}
\centering
\begin{subtable}[t]{0.48\textwidth}
\centering
\begin{tabular}{lrrrrr}
\toprule
\multicolumn{1}{c}{} & \multicolumn{2}{c}{\textsc{In-context}} & \multicolumn{2}{c}{\textsc{Lin. Seg.}} \\
\cmidrule(lr){1-1}\cmidrule(lr){2-3}\cmidrule(lr){4-5}
Epochs & VOC  & ADE20K  & VOC  & ADE20K \\
\midrule
1 & 76.6 & 35.2 & 89.4 &  46.1 \\
2 & 76.6 & 35.3 & 89.4 & 45.8 \\
3 & 76.7 & 35.3 & 89.4 & 46.0 \\
5 & \textbf{76.7} & \textbf{35.3} & \textbf{89.4} & \textbf{46.2} \\
7 & 76.6 & 35.2 & 89.4 & 45.9 \\
10 & 76.7 & 35.3 & 89.3 & 45.8 \\
\bottomrule
\end{tabular}
\caption{Number of Epochs on COCO Dataset}
\label{tab:ablations_rasa_epochs_left}
\end{subtable}
\begin{subtable}[t]{0.48\textwidth}
\centering
\begin{tabular}{lrrrrr}
\toprule
\multicolumn{1}{c}{} & \multicolumn{2}{c}{\textsc{In-context}} & \multicolumn{2}{c}{\textsc{Lin. Seg.}} \\
\cmidrule(lr){1-1}\cmidrule(lr){2-3}\cmidrule(lr){4-5}
Epochs & VOC  & ADE20K  & VOC  & ADE20K \\
\midrule
1 & 76.1 & 34.5 & 89.4 & 45.9 \\
2 & 76.3 & 34.6 & 89.3 & 46.2 \\
3 & 76.4 & 34.9 & 89.4 & 45.8 \\
5 & \textbf{76.7} & \textbf{35.3} & \textbf{89.4} & 46.0 \\
7 & 76.5 & 35.0 & 89.3 & 46.0 \\
10 & 76.6 & 35.0 & 89.3 & \textbf{46.1} \\
\bottomrule
\end{tabular}
\caption{Number of Epochs on Pascal VOC}
\label{tab:ablations_rasa_epochs_right}
\end{subtable}
\label{tab:ablations_rasa_epochs}
\end{table*}

We further ablate the effect of the number of training epochs used for each dual-linear position predictor in the RASA head. Results on both \textsc{COCO} and \textsc{Pascal VOC} are reported in Tables~\ref{tab:ablations_rasa_epochs_left} and \ref{tab:ablations_rasa_epochs_right}.  

\vspace{0.5em}
\noindent\textbf{Observations.}  
Across both datasets, performance improves gradually from 1 to 3 epochs, with the most consistent gains observed when training for $5$ epochs. On \textsc{COCO} in \autoref{tab:ablations_rasa_epochs_left}, $5$ epochs achieves the strongest overall performance ($76.7$ VOC / $35.3$ ADE20K for in-context, $89.4$ VOC / $46.2$ ADE20K for linear segmentation). Beyond this point, additional training ($7$ or $10$ epochs) does not yield further benefits and in some cases slightly reduces performance, particularly on ADE20K segmentation.  

A similar trend is evident when training on \textsc{Pascal VOC} in \autoref{tab:ablations_rasa_epochs_right}. The $5$-epoch configuration again provides the best balance across both metrics ($76.7$ VOC / $35.3$ ADE20K in-context, $89.4$ VOC / $46.0$ ADE20K linear segmentation). Extending to $7$ or $10$ epochs produces only marginal changes, with scores fluctuating around the same plateau. Interestingly, ADE20K segmentation shows a minor uptick at $10$ epochs ($46.1$), but the difference relative to $5$ epochs is negligible and within noise.  

\vspace{0.5em}
\noindent\textbf{Conclusion.}  
These results highlight that training the dual-linear position predictor is \emph{computationally efficient}, requiring only around $5$ epochs to converge. Extending beyond $5$ epochs yields no meaningful improvements and may even slightly degrade results. Moreover, the stability of the trends across both \textsc{COCO} (scene-centric) and \textsc{Pascal VOC} (object-centric) datasets reinforces the robustness of this setting. In practice, we adopt $5$ epochs as the default configuration for RASA training.

\subsubsection{Number of Iterations for Training the RASA Head}
\label{app:rasa_iterations}
\begin{table*}[t]
\caption{\emph{Ablating the number of dual predictor layers used for training RASA head (i.e., the number of iterations)}.}
\setlength{\tabcolsep}{3pt}
\centering
\begin{subtable}[t]{0.48\textwidth}
\centering
\begin{tabular}{lrrrrr}
\toprule
\multicolumn{1}{c}{} & \multicolumn{2}{c}{\textsc{In-context}} & \multicolumn{2}{c}{\textsc{Lin. Seg.}} \\
\cmidrule(lr){1-1}\cmidrule(lr){2-3}\cmidrule(lr){4-5}
Iters & VOC  & ADE20K  & VOC  & ADE20K \\
\midrule
1 & 76.2 & 34.2 & 89.3 & 46.1 \\
2 & 76.2 & 34.5 & 89.4 & \textbf{46.1} \\
3 & 76.4 & 34.5 & 89.3 & 45.9 \\
4 & 76.5 & 34.6 & 89.3 & 45.9 \\
5 & 76.3 & 34.5 & 89.3 & 46.0 \\
6 & 76.4 & 34.8 & 89.4 & 45.8 \\
8 & \textbf{76.7} & \textbf{35.3 }& \textbf{89.4 }& 46.0 \\
10 & 76.3 & 35.1 & 89.3 & 46.0 \\
\bottomrule
\end{tabular}
\caption{Number of Iterations with Pascal VOC}
\label{tab:ablations_rasa_iterations_left}
\end{subtable}
\begin{subtable}[t]{0.48\textwidth}
\centering
\begin{tabular}{lrrrrr}
\toprule
\multicolumn{1}{c}{} & \multicolumn{2}{c}{\textsc{In-context}} & \multicolumn{2}{c}{\textsc{Lin. Seg.}} \\
\cmidrule(lr){1-1}\cmidrule(lr){2-3}\cmidrule(lr){4-5}
Iters & VOC  & ADE20K  & VOC  & ADE20K \\
\midrule
1 & 76.1 & 34.2 & 89.3 & 45.9 \\
2 & 76.2 & 34.5 & 89.4 & 46.1 \\
3 & 76.3 & 34.5 & 89.4 & 46.0 \\
4 & 76.5 & 34.5 & 89.4 & 46.2 \\
5 & 76.3 & 34.5 & 89.4 &46.0 \\
6 & 76.4 & 34.8 & 89.4 &  46.0 \\
8 & \textbf{76.7} & \textbf{35.3} & \textbf{89.4} & \textbf{46.2} \\
10 & 76.6 & 35.2 & 89.4 & 46.0 \\
\bottomrule
\end{tabular}
\caption{Number of Iterations with Coco}
\label{tab:ablations_rasa_iterations_right}
\end{subtable}
\label{tab:ablations_rasa_iterations}
\end{table*}

Finally, we ablate the number of \emph{iterations} used to train the RASA head, where each iteration corresponds to training an additional dual-linear position predictor and integrating the previously trained predictors into the final ViT block. Results on \textsc{Pascal VOC} and \textsc{COCO} are reported in Tables~\ref{tab:ablations_rasa_iterations_left} and \ref{tab:ablations_rasa_iterations_right}.  

\vspace{0.5em}
\noindent\textbf{Observations.}  
Even a small number of iterations (1--2) produces competitive results across both datasets, suggesting that the RASA head can already provide strong positional disentanglement with minimal overhead. For example, with only 2 iterations we obtain $76.2$ VOC / $34.5$ ADE20K (in-context) and $89.4$ VOC / $46.1$ ADE20K (linear segmentation) on \textsc{Pascal VOC}, which is close to the final performance.  

However, performance steadily improves as the number of iterations increases, with the most notable gains occurring between 2 and 8 iterations. At $8$ iterations, results peak at $76.7$ VOC / $35.3$ ADE20K in-context and $89.4$ VOC / $46.0$ ADE20K for segmentation on \textsc{Pascal VOC} (in \autoref{tab:ablations_rasa_iterations_left}), and $76.7$ VOC / $35.3$ ADE20K in-context and $89.4$ VOC / $46.2$ ADE20K on \textsc{COCO} (in \autoref{tab:ablations_rasa_iterations_right}). Beyond this point ($10$ iterations), performance saturates and in some cases slightly declines, suggesting diminishing returns from further iterative training.  

Interestingly, across both datasets, the segmentation metrics appear to plateau earlier (around 4--6 iterations), while in-context segmentation continues to benefit until iteration 8. This indicates that additional iterations primarily refine segmentation-related representations, even though these could be sort of learned by the linear layer used in linear segmentation.  

\vspace{0.5em}
\noindent\textbf{Conclusion.}  
These experiments demonstrate that the RASA head achieves strong results even with a small number of iterations, but the optimal trade-off between performance and compute is obtained at $8$ iterations. Beyond this, additional predictors provide no meaningful improvements, confirming that the gains saturate after moderate iterative refinement.

\end{document}